\documentclass[letterpaper,11pt]{article}
\usepackage[margin=1in]{geometry}
\usepackage{algorithm,algorithmic}
\usepackage{graphicx}
\usepackage{amsmath,amsfonts,amssymb,theorem,euscript,array,enumerate,amsfonts,mathrsfs}
\usepackage{subcaption}
\usepackage{ulem}
\usepackage{bm, bbm}
\usepackage[colorlinks=true,breaklinks=true,bookmarks=true,urlcolor=blue,citecolor=blue,linkcolor=blue,bookmarksopen=false,draft=false]{hyperref}

\usepackage{endnotes}
\let\footnote=\endnote

\usepackage{natbib}
\bibpunct[, ]{[}{]}{,}{n}{}{,}%

\numberwithin{equation}{section}

\newtheorem{Theorem}{Theorem}%[part]
\newtheorem{Definition}{Definition}%[part]
\newtheorem{Assumption}{Assumption}%[part]
\newtheorem{Lemma}{Lemma}%[part]
\newtheorem{Remark}{Remark}%[part]
\usepackage[T1]{fontenc}
\usepackage[latin1]{inputenc}
\usepackage{hhline}
\usepackage{graphicx}
\usepackage{verbatim}
\usepackage{eurosym}
\usepackage{dsfont}

\def\esssup_#1{\underset{#1}{\mathrm{ess\,sup\, }}}
\def\essinf_#1{\underset{#1}{\mathrm{ess\,inf\, }}}
\def\argmax_#1{\underset{#1}{\mathrm{arg\,max\, }}}
\def\argmin_#1{\underset{#1}{\mathrm{arg\,min\, }}}

\def \A{\mathbb{A}}

\def \R{\mathbb{R}}

\def \E{\mathbb{E}}

\def \P{\mathbb{P}}

\def \A{{\cal A}}

\def \Cc{{\cal C}}

\def \Lc{{\cal L}}

\def \Oc{{\cal O}}
\def \Nc{{\cal N}}

\def \Xc{{\cal X}}

\def \beqs{\begin{eqnarray*}}
\def \enqs{\end{eqnarray*}}
\def \beq{\begin{eqnarray}}
\def \enq{\end{eqnarray}}

\begin{document}

\title{Adversarial Training for Gradient Descent: Analysis Through its Continuous-time Approximation}

\author{
Haotian Gu
\thanks{Department of Mathematics, UC Berkeley.
\textbf{Email:} 
haotian{\textunderscore}gu@berkeley.edu }
\and
Xin Guo
\thanks{Department of Industrial Engineering \& Operations Research, UC Berkeley.
\textbf{Email:} 
xinguo@berkeley.edu}
\and
Xinyu Li
\thanks{Department of Industrial Engineering \& Operations Research, UC Berkeley.
\textbf{Email:}
xinyu\_li@berkeley.edu}
}
\date{March 1, 2023}
\maketitle

\begin{abstract}
    Adversarial training has gained great popularity as  one of the most effective defenses for deep neural network and more generally for gradient-based machine learning models against adversarial perturbations on data points. This paper establishes a continuous-time approximation for  the mini-max game of adversarial training. This approximation approach allows for precise and analytical comparisons between stochastic gradient descent and its adversarial training counterpart; and confirms theoretically the robustness of adversarial training from a new gradient-flow viewpoint. The analysis is then corroborated through various analytical and  numerical examples. 
\end{abstract}

%%%%%%%%%%%%%%%%%%%%%%%%%%%%%%%%%%%%%%%%%%%
%%%%%%%%%%%%%%%%%%%%%%%%%%%%%%%%%%%%%%%%%%%

\section{Introduction}

\begin{figure}
    \centering
    \includegraphics[width=12cm]{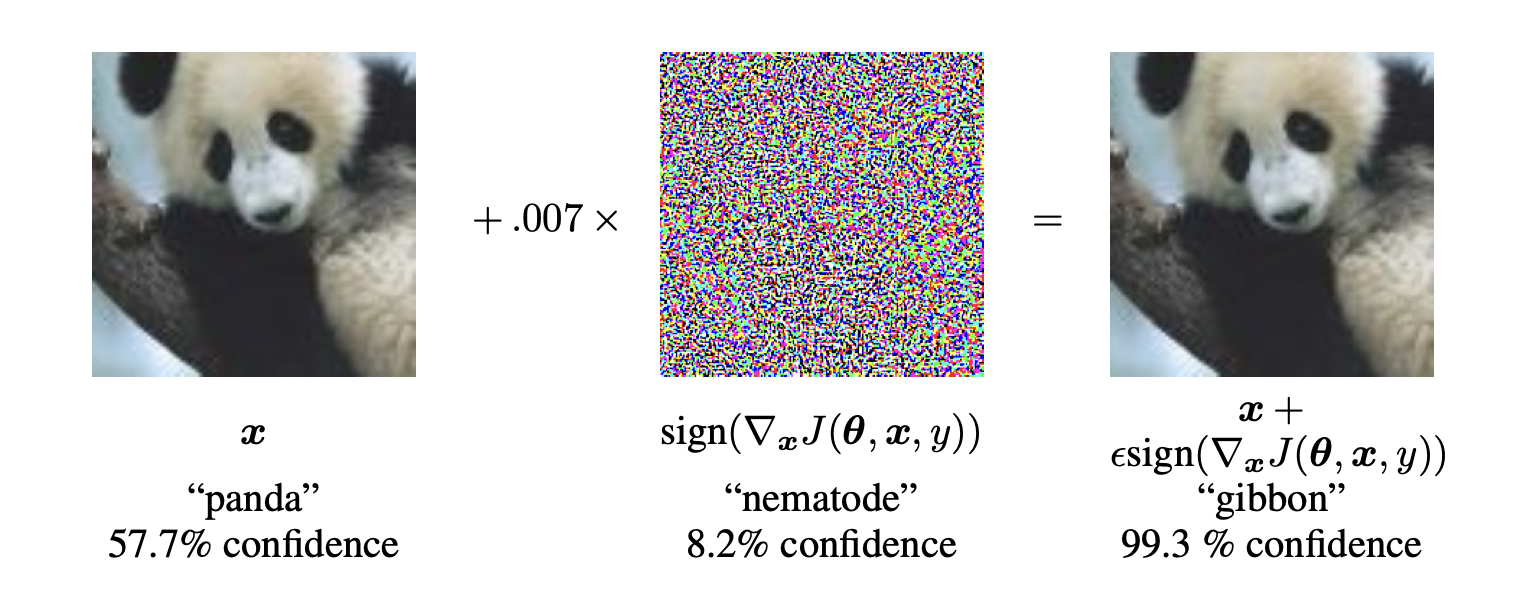}
    \caption{A demonstration of adversarial  generation applied to GoogleNet \cite{szegedy2013intriguing} on ImageNet. By adding an imperceptibly small vector whose elements are equal to the sign of the elements of the gradient of the cost function with respect to the input,  GoogLeNet's classification of the image is utterly changed. Here $\epsilon$ of $.007$ corresponds to the magnitude of the smallest bit of an $8$-bit image encoding after GoogLeNet's conversion to real numbers.}
    \label{fig:panda example}
\end{figure}

Deep neural networks and more generally gradient-based machine learning models have enjoyed substantial successes in many applications. Their performance, however, can significantly deteriorate by small and human imperceptible adversarial perturbations. Figure \ref{fig:panda example} illustrates such a well-known instance of adversarial attack which leads to blatantly identification errors \cite{szegedy2013intriguing, goodfellow2014explaining}.
Such vulnerability of machine learning models raises concerns of their practicability in robustness-critical applications. 

Adversarial training, proposed in \cite{madry2018towards}, is one of the most promising defenses for machine learning models against adversarial perturbation. Recent empirical studies such as \cite{carlini2017towards} and \cite{athalye2018obfuscated} have demonstrated its effectiveness and robustness of training neural networks in many applications.
The idea of adversarial training \cite{madry2018towards} is to formulate a mini-max game between a learner who aims to improve the model performance and an adversary who is allowed to perturb the data inputs. Algorithmically speaking, in each round, the adversary generates new adversarial examples against the current neural network via projected gradient descent (PGD), the learner then responds by taking a gradient step to decrease its loss. This procedure is prescribed in  Algorithm \ref{algo:Adv_general}.

Given the popularity of adversarial training, considerable effort has been made to further improve its performance. \cite{yan2018deepdefense} and \cite{farnia2018generalizable} use Lipschitz regularizations for better generalization performance of trained models; \cite{YOPO2019Zhang} designs a computationally efficient variant of adversarial training based on the Pontryagin Maximum Principle from robust controls; and \cite{zhang2020attack} suggests an early-stopped PGD to generate adversarial examples. To improve robustness, \cite{maini2020adver} extends the standard PGD procedure to incorporate multiple perturbation models into a single attack.

Parallel to these empirical successes, there are growing research interests in analyzing convergence and robustness  of adversarial training (\cite{madry2018towards}). For instance, \cite{wang2019convergence}
considers quantitatively evaluating the convergence quality of adversarial examples found by the adversary, in order to ensure the convergence and the robustness; \cite{gao2019convergence} and \cite{zhang2020over} study the  convergence and the robustness of adversarial training on over-parameterized neural networks; \cite{seidman2020robust} provides convergence analysis of adversarial training by combining techniques from robust optimal controls and inexact oracle methods from optimization; \cite{deng2020towards} investigates the non-concave landscape
of the adversary for a two-layer neural network with a quadratic loss function; \cite{yin2019rademacher} studies the adversarially robust generalization problem via Rademacher complexity; \cite{chen2020more} characterizes the generalization gap in terms of the number of training samples for Gaussian and Bernoulli models; and finally, \cite{raghunathan2020understand} explores the trade-off between the robustness and the accuracy in linear models.

Given these empirical and theoretical advances, it is natural to ask: what is the difference between vanilla stochastic gradient descent and adversarial training? Is it possible to analytically quantify this difference? And, how the hyper-parameters such as perturbation step and learning rate, affect the performance of adversarial training? 
These are the focuses of our study in this paper.

\paragraph{Our work} 
This paper considers the mini-max game of adversarial training by alternating stochastic gradient ascent and descent. By establishing a continuous-time approximation to the training process in the form of a stochastic differential equation (SDE), it enables  precise and analytical comparisons stochastic gradient descent and its adversarial training counterpart. Moreover, it confirms theoretically the robustness of adversarial training from a new gradient-flow viewpoint. This theoretical study is corroborated via several analytical and numerical examples. In the robust portfolio selection problem, it reveals an intriguing connection between adversarial training and the robust in portfolio selections, under appropriate choices of hyper-parameters such as the learning rate and the iteration steps. In the experiment with a logistic regression, it demonstrates that adversarial training leads to both increased robustness and reduced losses.  

\paragraph{Related works}
The idea of approximating discrete-time stochastic gradient algorithms (SGAs) by continuous-time SDE dynamics can be traced back to \cite{Mandt2015continuous} and \cite{Li2017Stochastic}. %In traditional convergence analysis for SGAs, different techniques must be adopted for different variants of the algorithms.
%In contrast, the SDE approximation framework provides  a more systematic approach to analyze dynamics of SGAs. 
It has recently been extended to various settings of SGAs. For instance, \cite{Krichene2017acceleration} and \cite{an2020stochastic} establish SDE approximations of accelerated mirror descent and asynchronous SGD, respectively; \cite{chaudhari2018deep} designs an entropy-regularized training algorithm motivated by the SDE approximation; \cite{chaudhari2018stochastic} builds the connection between SGD and variational inference by considering the evolution of training parameters; and finally, \cite{cao2020approximation} studies the training process of generative adversarial networks (GANs) via a coupled SDEs system.

The formulation of adversarial training is also closely related to robust optimization \cite{ben2009robust}, where the goal is to optimize the model's worst-case performance under data uncertainty. Robust optimization has a number of applications, including financial portfolio optimization \cite{blanchet2021distributionally}, \cite{data_driven},  statistics \cite{nguyen2020distributionally}, machine learning \cite{blanchet2019robust}) and reinforcement learning \cite{si2020distributional}). Recently, \cite{sinha2017certifying} and \cite{ren2022distributionally} investigate the empirical performance of solving robust optimization problem with adversarial training. Our study on the robust portfolio selection
problem shows an intriguing connection between robust optimization with adversarial training, under proper choices of hyper-parameters.

\paragraph{Organization}  Section \ref{sec:setting} presents the problem set-up for adversarial training. Section \ref{sec:approx} establishes the continuous-time SDE approximation for adversarial training, with error bound and robustness analysis, as well as discussion regarding the convergence of adversarial training via the invariant measure of the SDE. Section \ref{sec:SGD_VS_AL} compares the vanilla stochastic gradient descent with adversarial training from the continuous-time SDE viewpoint. Section \ref{sec:RO_vs_AT} draws the connection between robust optimization and adversarial training through a robust portfolio selection problem. Section \ref{sec:proof} is devoted to the technical proofs of main convergence results presented in Section \ref{sec:approx}.
Finally, Section \ref{sec:numeric} illustrates the theoretical results in the previous sections with numerical experiments. Meanwhile, the robust portfolio optimization problem in Section \ref{sec:RO_vs_AT} is solved numerically using adversarial training, and the impacts of hyper-parameters will be discussed with theoretical explanations. 

\paragraph{Notations}
The following notations will be used throughout the appendix.
\begin{itemize}
    \item For $p \geq 1,\|\cdot\|_{p}$ denotes the $p$-norm over $\mathbb{R}^{d},$ i.e., $\|x\|_{p}=\left(\sum_{i=1}^{d}\left|x_{i}\right|^{p}\right)^{\frac{1}{p}}$ for any $x \in \mathbb{R}^{d}$. 
    
    \item Let $M$ by a $n$-by-$m$ real-valued matrix with the $i,j$-th entry $m_{ij}$. $\|M\|_2=\sqrt{\sum_{i,j}m_{ij}^2}$ denotes the Frobenius norm of $M$. $M^T$ denotes the transpose of $M$.
    
    \item Let $\mathcal{X}$ be an arbitrary nonempty subset of $\mathbb{R}^{d_1},$ and $f$ is a function from $\Xc$ to $\R^{d_2}$. We say $f$ is Lipschitz continuous if there exists some constant $L>0$, such that for any $x,y\in\Xc$,
    $$\|f(x)-f(y)\|_2\leq L\|x-y\|_2.$$
    
    \item Let $\mathcal{X}$ be an arbitrary nonempty subset of $\mathbb{R}^{d},$ the set of $k$ continuously differentiable functions over some domain $\mathcal{X}$ is denoted by $\mathcal{C}^{k}(\mathcal{X})$ for any nonnegative integer $k.$ In particular when $k=0, \mathcal{C}^{0}(\mathcal{X})=\mathcal{C}(\mathcal{X})$ denotes the set of continuous functions.
    
    \item Let $J=\left(J_{1}, \ldots, J_{d}\right)$ be a $d$-tuple multi-index of order $|J|=\sum_{i=1}^{d} J_{i},$ where $J_{i}$ is a nonnegative integer for all $i=1, \ldots, d;$ then define the operator $\nabla^{J}=\left(\partial_{1}^{J_{1}}, \ldots, \partial_{d}^{J_{d}}\right)$.
    
    \item Fix an arbitrary $\alpha \in \mathbb{Z}^{+}$. $\mathcal{G}^{\alpha}\left(\mathbb{R}^{d}\right)$ denotes a subspace of $\mathcal{C}^{\alpha}\left(\mathbb{R}^{d}\right)$, where for any $g \in \mathcal{G}^{\alpha}\left(\mathbb{R}^{d}\right)$ and any $d$-tuple multi-index $J$ with $|J| \leq \alpha,$ there exist $k_{1}, k_{2} \in \mathbb{N}$ such that 
    $$\nabla^{J} g(x) \leq k_{1}\left(1+\|x\|_{2}^{2 k_{2}}\right), \quad \forall x \in \mathbb{R}^{d},$$
    i.e. $g$'s partial derivatives up to and including order $\alpha$ have at most polynomial growth. %In particular, $G$ denotes the space of continuous functions with at most polynomial growth.
    
    \item The Wasserstein distance between two probability measures $\mathbb{Q}_1, \mathbb{Q}_2$ are
    \begin{align*}
    \mathcal{W}_c(\mathbb{Q}_1, \mathbb{Q}_2) := \inf_{\Pi} \{& \int c(\xi_1, \xi_2)\Pi(d\xi_1, d\xi_2), \Pi \text{ is a joint distribution of } \\ 
    & \xi_1 \text{ and } \xi_2 \text{ with marginals } \mathbb{Q}_1 \text{ and } \mathbb{Q}_2,\}
    \end{align*}
    
    \item For any open subset $\mathcal{O}$ in $\mathbb{R}^d$,
    $\Cc(\mathcal{O})$ denotes all real-valued continuous functions on $\mathcal{O}$,
    $\Cc^{k}(\mathcal{O})$ denotes all k-times continuously differentiable functions on $\mathcal{O}$. Given $t_1 < t_2$, for any $Q = [t_1, t_2) \times \mathcal{O},$ $\Cc^{k,j}(Q)$ contains functions $g:(t,x)\in Q \rightarrow \mathbb{R}^d $ whose partial derivatives of orders $\leq k$ in $t$ and orders $\leq j$ in $x$ are continuous.
    
    \item $\partial \mathcal{O}$ is defined to be of class $\Cc^{(k)}$ if for all $z \in \partial  \mathcal{O}$, there exists a radius $r=r\left(z\right)>0$ such that, up to relabeling the variables, $B_r\left(z\right) \cap  \mathcal{O}=\left\{x \in B_r\left(z\right): x_d>\gamma\left(x_1, \ldots, x_{d-1}\right)\right\}$ 
    for some $\Cc^k$ function $\gamma=\gamma\left(x_1, \ldots, x_{d-1}\right)$ on $B_r(z) \cap\left(\mathbb{R}^{d-1} \times\left\{
   z_d\right\}\right)$.
    
\end{itemize}

%%%%%%%%%%%%%%%%%%%%%%%%%%%%%%%%%%%%%%%%%%%
%%%%%%%%%%%%%%%%%%%%%%%%%%%%%%%%%%%%%%%%%%%

\section{Problem Setting}\label{sec:setting}
Given a data set $\{x_i\}_{i=i}^N\subset\mathbb{R}^d$ with $d, N\in\mathbb{Z}^+$, and a constraint set $\Delta\in\mathbb{R}^d$, the adversarial training problem is to find an appropriate model parameter $\theta \in\mathbb{R}^{d_\theta}$ and small perturbations $\delta_i$ to solve the following min-max optimization problem:
\begin{equation}\label{equ:adv_objective}
    \min_{\theta\in\mathbb{R}^{d_\theta}}\max_{\{\delta_i\}_{i=1}^N\subset\Delta}\frac{1}{N}\sum_{i=1}^NL(\theta, x_i+\delta_i).
\end{equation}
Here $L(\cdot, \cdot):\R^{d_\theta}\times\R^d\to\R$ is a loss function depending on both model parameter $\theta \in\mathbb{R}^{d_\theta}$ and data $x\in\R^d$, with $d_{\theta}$  the dimension of a given parameter space for $\theta$. Meanwhile, $\{\delta_i\}_{i=1}^N\subset\Delta$ is the perturbation on the data point $\{x_i\}_{i=1}^N$ within the constraint set $\Delta$. A common choice for $\Delta$ is $\{\delta_i\in\mathbb{R}^d: \|\delta_i\|_p\leq \epsilon\}$ with some given $\epsilon>0$ and $p=1,2$ or $\infty$.

One of the established approaches to compute \eqref{equ:adv_objective} is by performing a gradient ascent on the perturbation parameter $\delta$ and a gradient descent on the model parameter $\theta$. Such an alternating optimization algorithm, called projected-gradient-descent (PGD) adversarial training \cite{madry2018towards}, is shown in Algorithm \ref{algo:Adv_general}:  in the inner loop, the most powerful adversarial attack $\delta$ to the input data batch is obtained via the multi-step projected gradient ascent; and in the outer loop, $\theta$ is updated by the one-step gradient descent, based on the perturbed batch of data.

\begin{algorithm}[!ht]
  \caption{\textbf{Projected Gradient Descent Adversarial Training \cite{madry2018towards}}}
  \label{algo:Adv_general}
\begin{algorithmic}[1]
    \STATE \textbf{Input}: loss function $L$, training set $\{x_i\}_{i=1}^N$, mini-batch size $B$, training step $T$, PGD step $K$, learning rates for outer and inner loops $\eta_{O}, \eta_{I}$, perturbation constraint set $\Delta$.
    \STATE \textbf{Initialize}: $\theta_0$.
    \FOR {$ 1 \leq t \leq T$}
        \STATE Sample a mini-batch of size $B$: $\{x_{i_1},\dots,x_{i_B}\}$. 
        \STATE Set $\delta_0=0$, $\widehat{x}_j=x_{i_j},j=1,\dots,B$.
        \FOR {$ 1 \leq k \leq K$}
            \STATE $\delta_k=\Pi_\Delta(\delta_{k-1}+\frac{\eta_{I}}{B}\sum_{j=1}^B\nabla_xL(\theta_{t-1},\widehat{x}_j+\delta_{k-1}))$
        \ENDFOR
        \STATE $\theta_t=\theta_{t-1}-\frac{\eta_O}{B}\sum_{j=1}^B\nabla_\theta L(\theta_{t-1},\widehat{x}_j+\delta_K))$
    \ENDFOR.
\end{algorithmic}
\end{algorithm}

Our goal is to analyze the analytical impact of adversarial perturbations. 
Here, we consider a more general form that incorporate variants of the original min-max problem \eqref{equ:adv_objective}: 
\begin{equation}\label{equ:adv_objective_new}
\min_{\theta\in\mathbb{R}^{d_\theta}}\max_{\{\delta_i\}_{i=1}^N}J(\theta, x, \delta)
=:\min_{\theta\in\mathbb{R}^{d_\theta}}\max_{\{\delta_i\}_{i=1}^N}\frac{1}{N}\sum_{i=1}^NL(\theta, x_i+\delta_i)-\lambda\cdot R(\delta_i),
\end{equation}
where the function $R:\mathbb{R}^d\to\R$ is the regularization term with $\lambda$ a hyper-parameter. For instance, when the original constraint set $\Delta$ is in the form $\{\delta_i\in\mathbb{R}^d: R(\delta_i)\leq 0\}$, then the modified problem \eqref{equ:adv_objective_new} is a Lagrange relaxation  of the original problem \eqref{equ:adv_objective}. 
Moreover, $R$ is assumed to be a convex function attaining the minimum at the origin, with $\nabla_\delta R(0)=0$. This is consistent with the literature for adversarial training, including the popular  $l_p$ regularization.

%\begin{equation}\label{equ:new_J}
%    J(\theta, x, \delta) = L(\theta, x+\delta)-\lambda\cdot R(\delta).
%\end{equation}

\begin{algorithm}[!ht]
  \caption{\textbf{Adversarial Training with Modified Objective \eqref{equ:adv_objective_new}}}
  \label{algo:Adv_relax}
\begin{algorithmic}[1]
    \STATE \textbf{Input}: loss function $J(\theta, x, \delta) = L(\theta, x+\delta)-\lambda R(\delta)$, training set $\{x_i\}_{i=1}^N$, mini-batch size $B$, training step $T$, inner loop step $K$, learning rates for outer and inner loops $\eta_{O}, \eta_{I}$.
    \STATE \textbf{Initialize}: $\theta_0$.
    \FOR {$ 1 \leq t \leq T$}
        \STATE Sample a mini-batch of size $B$: $\{x_{i_1},\dots,x_{i_B}\}$. 
        \STATE Set $\delta_0=0$, $\widehat{x}_j=x_{i_j},j=1,\dots,B$.
        \FOR {$ 1 \leq k \leq K$}
            \STATE $\delta_k=\delta_{k-1}+\frac{\eta_{I}}{B}\sum_{j=1}^B\nabla_\delta J(\theta_{t-1},\widehat{x}_j,\delta_{k-1})$
        \ENDFOR
        \STATE $\theta_t=\theta_{t-1}-\frac{\eta_O}{B}\sum_{j=1}^B\nabla_\theta J(\theta_{t-1},\widehat{x}_j,\delta_K))$
    \ENDFOR.
\end{algorithmic}
\end{algorithm}

Our approach is to  establish a continuous-time approximation for the discrete-time Algorithm \ref{algo:Adv_relax}, with analysis of the approximation error bound. This approximation will enable us to  compare analytically the PGD with adversarial training versus its vanilla form, as well as the robustness of adversarial training. 
%%%%%%%%%%%%%%%%%%%%%%%%%%%%%%%%%%%%%%%%%%%
%%%%%%%%%%%%%%%%%%%%%%%%%%%%%%%%%%%%%%%%%%%

\section{Continuous-time Approximation}\label{sec:approx}
To establish the continuous-time SDE dynamic for adversarial training, 
let us first introduce some notations. 

\paragraph{Notation} Let $\P_x$ be the probability distribution of data $x$ on $\R^d$. Throughout the paper, if there is no additional specification, we assume the expectation $\E[\cdot]$ is taken over the distribution $\P_x$. Let $L(\cdot, \cdot):\R^{d_\theta}\times\R^d\to\R$ be the loss function depending on both model parameter $\theta \in\mathbb{R}^{d_\theta}$ and data $x\in\R^d$. Define $\nabla_x L$ and $\nabla_\theta L$ to be the gradients of $L$ with respect to $x$ and $\theta$, respectively. Define $\nabla_{x\theta}L$ to be the matrix whose $i,j$-th entry is $\frac{\partial^2L}{\partial x_j\partial\theta_i}$.

\paragraph{Assumptions} 
Throughout this paper, 
we will assume for ease of exposition and with little loss of generality the same constant learning rate $\eta$ for both the inner and outer loops, In addition, we assume:

\begin{Assumption}\label{ass:L}
The loss function $L$ satisfies the following conditions.
\begin{enumerate}
    \item $L\in\Cc^3(\R^{d_\theta+d})$.
    
    \item For any $\theta\in\R^{d_\theta}$, $\E[|L(\theta,x)|]<\infty$ and $\E[\|\nabla_xL(\theta,x)\|_2^2]<\infty$.
    
    \item For any $R>0$, there exists $M_{R}>0$ such that $\P_x$-almost surely, 
    $$\max_{\|\theta\|_2\leq R}\max\left\{\|\nabla_x L(\theta,x)\|_2, \|\nabla_{x\theta} L(\theta,x)\|_2, \|\nabla_{x\theta} L(\theta,x)\nabla_x L(\theta,x)\|_2\right\}\leq M_R.$$
\end{enumerate}
\end{Assumption}

Note that in the empirical risk minimization case \cite{vapnik1991principles}, where the support of $\P_x$ is a finite set, the conditions above are often trivially satisfied. Meanwhile, Assumption \ref{ass:L}.2 allows us to define the following functions
\begin{equation}\label{equ:g&D}
    g(\theta):=\E[L(\theta,x)],\quad I(\theta):=\E[\nabla_xL(\theta,x)],\quad H(\theta):=\E[\|\nabla_xL(\theta,x)\|^2_2],
\end{equation}
\begin{equation}\label{equ:var_x}
    \Sigma(\theta):=\text{Var}_x\left(\nabla_\theta L(\theta,x)\right).
\end{equation}
Moreover, Assumption \ref{ass:L} and the dominated convergence theorem allows changing orders of  expectations and derivatives such that
\begin{align*}
    &\nabla_\theta g(\theta)=\nabla_\theta\E[L(\theta,x)]=\E[\nabla_\theta L(\theta,x)],\\
    &\nabla_\theta I(\theta)=\nabla_\theta\E[\nabla_x L(\theta,x)]=(\E[\nabla_{x\theta} L(\theta,x)])^T,\\
    &\nabla_\theta H(\theta)=\nabla_\theta\E[\|\nabla_xL(\theta,x)\|^2_2]=\E[\nabla_\theta(\|\nabla_xL(\theta,x)\|^2_2)]=2\E[\nabla_{x\theta} L(\theta,x)\nabla_x L(\theta,x)].
\end{align*}
Hence, under Assumption \ref{ass:L}, the functions $g(\theta)$, $I(\theta)$ and $H(\theta)$ are differentiable.

The approximation is established through a series of analysis.
\subsection{Firs Step: One-step Difference in Discrete-time Update} 
We first focus on analyzing one step in Algorithm \ref{algo:Adv_relax} to see how the parameter $\theta$ is updated from $t$ to ${t+1}$. 

Consider an inner loop starting with $\delta_0=0$, then
\begin{align*}
    \delta_1&=\delta_0+\frac{\eta}{B}\sum_{j=1}^B\nabla_\delta J(\theta_{t},\widehat{x}_j,0)=\frac{\eta}{B}\sum_{j=1}^B\nabla_x L(\theta_{t},\widehat{x}_j),\\
    \delta_2&=\delta_1+\frac{\eta}{B}\sum_{j=1}^B\left(\nabla_x L(\theta_{t},\widehat{x}_j+\delta_1)-\lambda\nabla_\delta R(\delta_1)\right)\\
    &=\frac{2\eta}{B}\sum_{j=1}^B\nabla_x L(\theta_{t},\widehat{x}_j)+\Oc(\eta^2).
\end{align*}
The second equality comes from Taylor's expansion at $\delta=0$. Continuing this calculation, we see that any $K\ge 2$, 
\begin{equation}\label{equ:inner_loop_delta}
    \delta_K=\frac{K\eta}{B}\sum_{j=1}^B\nabla_x L(\theta_{t},\widehat{x}_j)+\Oc(\eta^2).
\end{equation}
Since we only keep the first order terms,  terms involving higher order derivatives of $R$ are negligible by the assumption of $R$.

Given \eqref{equ:inner_loop_delta}, by Assumption \ref{ass:L} and the third order Taylor's expansion, the update on $\theta$ from the outer loop is given by
\begin{align}\label{equ:one_step_theta}
    \theta_{t+1}=&\,\theta_t-\frac{\eta}{B}\sum_{j=1}^B\nabla_\theta L\left(\theta_{t},\widehat{x}_j+\frac{K\eta}{B}\sum_{j=1}^B\nabla_x L(\theta_{t},\widehat{x}_j)+\Oc(\eta^2)\right)\notag\\
    =&\,\theta_t-\frac{\eta}{B}\sum_{j=1}^B\nabla_\theta L\left(\theta_{t},\widehat{x}_j\right)-\frac{K\eta^2}{B^2}\sum_{i,j=1}^B\nabla_{x\theta} L(\theta_{t},\widehat{x}_j)\nabla_x L(\theta_t,\widehat{x}_i)+\Oc(\eta^3).
\end{align}
  In particular, given an initial model parameter $\theta_0$ and independent samples  $\{\widehat{x}_j\}_{j=1}^B$  from $\P_x$, Algorithm \ref{algo:Adv_relax} updates $\theta_1$ as
\begin{align}\label{equ:one_step_theta_0}
    \theta_{1}=&\,\theta_0-\frac{\eta}{B}\sum_{j=1}^B\nabla_\theta L\left(\theta_{0},\widehat{x}_j+\frac{K\eta}{B}\sum_{j=1}^B\nabla_x L(\theta_{0},\widehat{x}_j)+\Oc(\eta^2)\right)\notag\\
    =&\,\theta_0-\frac{\eta}{B}\sum_{j=1}^B\nabla_\theta L\left(\theta_{0},\widehat{x}_j\right)-\frac{K\eta^2}{B^2}\sum_{i,j=1}^B\nabla_{x\theta} L(\theta_{0},\widehat{x}_j)\nabla_x L(\theta_0,\widehat{x}_i)+\Oc(\eta^3).
\end{align}

Now, in order to find a continuous-time approximation for $\theta_t$, key quantities are the first and second order moments $\E[D]$ and $\E[DD^T]$,
with $D:=\theta_{1}-\theta_0$ the one-step difference of $\theta$. The expectation here is taken over the randomness of the mini-batch samples $\{\widehat{x}_j\}_{j=1}^B$. By direct computations,
\begin{align*}
\begin{split}
    \E[D]&=\,-\frac{\eta}{B}\sum_{j=1}^B\E[\nabla_\theta L\left(\theta_{0},\widehat{x}_j\right)]-\frac{K\eta^2}{B^2}\sum_{i,j=1}^B\E[\nabla_{x\theta} L(\theta_{0},\widehat{x}_j)\nabla_x L(\theta_0,\widehat{x}_i)]+\Oc(\eta^3)\\
    &=\,-\eta\E[\nabla_\theta L\left(\theta_{0},x\right)]-\frac{K\eta^2}{B^2}\sum_{i,j=1}^B\E[\nabla_{x\theta} L(\theta_{0},\widehat{x}_j)\nabla_x L(\theta_0,\widehat{x}_i)]+\Oc(\eta^3)\\
    &=\,-\eta\E[\nabla_\theta L(\theta_0,x)] - K\eta^2\E[\nabla_{x\theta} L(\theta_{0},x)]\E[\nabla_x L(\theta_0,x)] \\
    &\quad\,\,-\frac{K\eta^2}{B}\bigg(\E[\nabla_{x\theta} L(\theta_{0},x)\nabla_x L(\theta_0,x)] - \E[\nabla_{x\theta} L(\theta_{0},x)]\E[\nabla_x L(\theta_0,x)]\bigg)+\Oc(\eta^3)\\
    &=\,-\eta\nabla_\theta g(\theta_0) - \frac{K\eta^2}{2}\nabla_\theta(\|I(\theta_0)\|^2_2) - \frac{K\eta^2}{2B}\left(\nabla_\theta H(\theta_0) - \nabla_\theta(\|I(\theta_0)\|^2_2)\right)+\Oc(\eta^3).
\end{split}\\
\begin{split}
    \E[DD^T]&=\,\frac{\eta^2}{B^2}\sum_{i,j=1}^B\E[\nabla_\theta L\left(\theta_{0},\widehat{x}_i\right) \nabla_\theta L\left(\theta_{0},\widehat{x}_j\right)^T]
    +\Oc(\eta^3)\\
    &=\,\eta^2\E[\nabla_\theta L(\theta_0,x)]\E[\nabla_\theta L(\theta_0,x)]^T\\
    &\quad\,\,+\frac{\eta^2}{B^2}\sum_{i=1}^B\left(\E[\nabla_\theta L(\theta_0,\widehat{x}_i) \nabla_\theta L(\theta_0,\widehat{x}_i)^T] - \E[\nabla_\theta L(\theta_0,\widehat{x}_i)]\E[\nabla_\theta L(\theta_0,\widehat{x}_i)]^T\right) + \Oc(\eta^3)\\
    &=\,\eta^2\nabla_\theta g(\theta_0) \nabla_\theta g(\theta_0)^T+\frac{\eta^2}{B}\Sigma(\theta_0)+\Oc(\eta^3).
\end{split}
\end{align*}
The formal computation leads to the following lemma.
\begin{Lemma}\label{lemma:moment_D}
    Assume Assumption \ref{ass:L}. Given an initial model parameter $\theta_0$, and assume that  $\{\widehat{x}_j\}_{j=1}^B$ are sampled independently from $\P_x$. Let $\theta_1$ be the one-step update of Algorithm \ref{algo:Adv_relax}, then the one-step difference $D=\theta_1-\theta_0$ satisfies
    \begin{equation}\label{equ:first_moment_d}
        \E[D] = -\eta\nabla_\theta g(\theta_0) - \frac{K\eta^2}{2}\nabla_\theta(\|I(\theta_0)\|^2_2) - \frac{K\eta^2}{2B}\left(\nabla_\theta H(\theta_0) - \nabla_\theta(\|I(\theta_0)\|^2_2)\right)+\Oc(\eta^3),
    \end{equation}
    \begin{equation}\label{equ:second_moment_d}
        \E[DD^T] = \eta^2\nabla_\theta g(\theta_0) \nabla_\theta g(\theta_0)^T+\frac{\eta^2}{B}\Sigma(\theta_0)+\Oc(\eta^3).
    \end{equation}
    and for any $i,j,k\in\{1,\cdots,d_\theta\}$, $\E[D_i D_j D_k]=\Oc(\eta^3)$, where $D_i$ denotes the $i^\text{th}$ coordinate of $D$. 
\end{Lemma}

\subsection{Second Step: One-step Difference in Continuous-time Approximation}
Next, we will find a continuous-time stochastic process $\{\Theta\}_{t\geq0}$ to approximate the discrete-time adversarial training dynamic \eqref{equ:one_step_theta}.

To this end, define $\{\Theta\}_{t\geq0}$ according to  the following stochastic differential equation:
\begin{equation}\label{equ:sde_dynamics}
    d\Theta_t=(b_0(\Theta_t)+\eta b_1(\Theta_t))dt+\sigma(\Theta_t)dW_t.
\end{equation} 
Here $\{W_t\}_{t\geq0}$ is a $d_{\theta}-$dimensional Brownian motion defined on a probability space $(\Omega,\mathcal{F},\mathbb{P})$, and the drift terms $b_0, b_1:\mathbb{R}^{d_\theta}\to\mathbb{R}^{d_\theta}$ and 
the diffusion term $\sigma:\mathbb{R}^{d_\theta}\to\mathbb{R}^{d_\theta}\times\mathbb{R}^{d_\theta}$ in \eqref{equ:sde_dynamics} are defined as:
\begin{align}
\begin{split}\label{equ:G}
    G(\theta)&:=g(\theta)+K\beta^{-1}(H(\theta) - \|I(\theta)\|^2_2),
\end{split}\\
\begin{split}\label{equ:b0}
    b_0(\theta)&:=-\nabla_\theta G(\theta),
\end{split}\\
\begin{split}\label{equ:b1}
    b_1(\theta)&:=-\frac{K}{2}\nabla_\theta(\|I(\theta)\|^2_2)-\frac{1}{4}\nabla_\theta(\|\nabla_\theta G(\theta)\|^2_2),
\end{split}\\
\begin{split}\label{equ:sigma}
    \sigma(\theta)&:=\sqrt{2\beta^{-1}}\Sigma(\theta)^{1/2}.
\end{split}
\end{align}
Here $\beta=2B/\eta$, the ratio between the batch size $B$ and the learning rate $\eta$, determines the scale of the diffusion. 
To ensure the well-definedness of the stochastic differential equation, we impose additionally the following smoothness conditions on $b_0(\theta)$, $b_1(\theta)$ and $\sigma(\theta)$.
\begin{Assumption}\label{ass:more_L}
$b_0,b_1$ and $\sigma$ defined in \eqref{equ:b0}, \eqref{equ:b1} and \eqref{equ:sigma} satisfy the following regularity conditions.
\begin{enumerate}
    \item $b_0,b_1$ and $\sigma$ are Lipschitz continuous.
    
    \item $b_0,b_1$ and $\sigma$ are in $\mathcal{G}^3(\R^{d_\theta})$.
\end{enumerate}
\end{Assumption}

Assumption \ref{ass:more_L}.1 is a standard assumption to guarantee that the SDE \eqref{equ:sde_dynamics} has a unique (strong) solution. Meanwhile, Assumption \ref{ass:more_L}.2 helps to control the growth of the solution (and its partial derivatives) of \eqref{equ:sde_dynamics}, which is crucial to the subsequent analysis of its approximation error.  These assumptions are satisfied for a neural network with sufficiently smooth activation and loss functions.

As one can see, $b_0,b_1$ and $\sigma$ defined in \eqref{equ:b0}, \eqref{equ:b1} and \eqref{equ:sigma} are appropriately chosen so that 
the dynamic \eqref{equ:sde_dynamics} with $b_0,b_1,\sigma$ defined in \eqref{equ:b0}-\eqref{equ:sigma} indeed matches the first and second moments of the discrete-time adversarial training dynamic \eqref{equ:one_step_theta}, as shown in the following lemma.

\begin{Lemma}\label{lemma:moment_C}
    Assume Assumption \ref{ass:L}, and Assumption \ref{ass:more_L}. The SDE dynamic \eqref{equ:sde_dynamics} with $\Theta_0=\theta_0$ admits a unique solution $\{\Theta_t\}_{t\geq0}$. Moreover, $\widetilde{D}:={\Theta}_\eta-\Theta_0$ satisfies
    $$\E[\widetilde{D}]=\E[D],\quad\E[\widetilde{D}\widetilde{D}^T]=\E[DD^T],$$
    where $\E[D]$ and $\E[DD^T]$ are defined in \eqref{equ:first_moment_d} and \eqref{equ:second_moment_d}, respectively.
    Moreover, for any $i,j,k\in\{1,\cdots,d_\theta\}$, $\E[\widetilde{D}_i\widetilde{D}_j\widetilde{D}_k]=\Oc(\eta^3)$, where $\widetilde{D}_i$ denotes the $i^\text{th}$ coordinate of $\widetilde{D}$.
\end{Lemma}

Now, we will continue the  error bound analysis of this approximation of $\theta$ by $\Theta$.

%%%%%%%%%%%%%%%%%%%%%%%%%%%%%%%%%%%%%%%%%%%

\subsection{Step Three:  Error Bound of Approximation }\label{subsec:approx_error}
We now provide a  rigorous proof that the SDE \eqref{equ:sde_dynamics} is indeed the continuous-time approximation for Algorithm \ref{algo:Adv_relax}, in some appropriate analytical sense:
since we are comparing a discrete-time stochastic process $\{\theta_t\}_{t=0,1,\cdot}$ with a continuous-time stochastic process $\{\Theta_t\}_{t\geq 0}$,
it is necessary to define an appropriate notion of approximation. 

Notice that the discrete-time process $\{\theta_t\}_{t=0,1,\cdots,N}$ is adapted to the filtration generated by $\left\{\widehat{x}_j\right\}_{j=1}^B$, which is by random sampling from the mini-batch, whereas the process $\{\Theta_t\}_{0\leq t\leq T}$ is adapted to a filtration generated by an independent $\{W_s\}_{0\leq s\leq t}$. Hence, it is not appropriate to compare individual sample paths. Instead,  we adopt the notion of \textit{weak approximation} to compare the distributions of sample paths instead of the sample paths themselves, following  \cite{li2019stochastic}. 
Recall
\begin{Definition}{(Order-$\alpha$ Weak Approximation)}
    Let $T>0, \eta \in(0,1 \wedge T),$ and $\alpha \geq 1$ be an integer. Set $N=\lfloor T / \eta\rfloor.$  A continuous-time stochastic process $\left\{X_{t}: t \in[0, T]\right\}$ is said of be an order-$\alpha$ weak approximation of a discrete stochastic process $\left\{x_{t}: t=0, \ldots, N\right\}$, if for every $g \in \mathcal{G}^{\alpha+1},$ there exists a positive constant $C,$ independent of $\eta,$ such that
    $$\max _{t=0, \cdots, N}\left|\mathbb{E} g\left(x_{t}\right)-\mathbb{E} g\left(X_{t \eta}\right)\right| \leq C \eta^{\alpha}.$$
\end{Definition}

The weak approximation is different from the strong approximation, where the actual sample-paths of  two processes are required to be close, for example,
$$
\max_{t=0,\cdots,N}\left\{\mathbb{E}\left[\left|x_{t}-X_{t \eta}\right|^{2}\right]\right\}\leq C \eta^{\alpha}.
$$
In contrast, weak approximation requires that the expectations of the two processes $\left\{X_{t}\right\}_{t\in[0, T]}$ and $\left\{x_{t}\right\}_{t=0,\ldots,N}$ over a sufficiently large class of test functions to be close. The test function class $\mathcal{G}^{\alpha+1}$ here includes all polynomials. In particular, it implies all moments of the two processes are close at the order of $\Oc(\eta^{\alpha})$.

As pointed out in \cite{li2019stochastic}, one important advantage of the weak approximation is that a continuous-time process can in fact approximate a discrete-time stochastic process whose step-wise driving noise is not Gaussian, as long as appropriate moments are matched (see the following Lemma \ref{lemma:one_to_n}). This additional flexibility is useful as it allows the treatment of more general classes of stochastic gradient iterations, for example, asynchronous stochastic gradient descent in \cite{an2020stochastic} and GAN training in \cite{cao2020approximation}.

The following lemma from \cite{li2019stochastic} is key to  show that the SDE \eqref{equ:sde_dynamics} is a second-order weak approximation for adversarial training dynamic \eqref{equ:one_step_theta}.

\begin{Lemma}{\cite{li2019stochastic}}\label{lemma:one_to_n}
Let $T>0, \eta \in(0,1 \wedge T)$ and $N=\lfloor T / \eta\rfloor .$ Let $\alpha \geq 1$ be an integer.
Let $\left\{X_{t}: t \in[0, T]\right\}$ be a continuous-time stochastic process satisfying
\begin{equation}\label{equ:general_contin}
    dX_t=b(X_t)dt+\sigma(X_t)dW_t.
\end{equation}
Here $\{W_t\}_{t\geq0}$ is a $d_{\theta}-$dimensional Brownian motion, and assume $b(x)$, $\sigma(x)$ are both Lipschitz continuous. Let $\left\{x_{t}: t=0, \cdots, N\right\}$ be a discrete-time stochastic process following
\begin{equation}\label{equ:general_dis}
    x_{t+1}=x_t + \eta h(x_t, y_t, \eta),
\end{equation}
where $h:\R^{d_1}\times\R^{d_2}\times\R\to\R^{d_1}$ and $\left\{y_{t}: t=0, \cdots, N\right\}\subset \R^{d_2}$ is a set of independent samples from some distribution $Y$, and $\left\{y_{t}: t=0, \cdots, N\right\}$ is independent of $\{W_t\}_{t\geq0}$. 

Define $\left\{X^s_{t}: t \in[0, T], X^s_0=s\right\}$ as the continuous-time stochastic process \eqref{equ:general_contin} starting from $s$, and $\left\{x^s_{t}: t=0, \cdots, N, x^s_{0}=s\right\}$ as the discrete-time process \eqref{equ:general_dis} starting from $s$.
Define $D(s)=x^s_1-s$ and $\widetilde{D}(s)=X^s_\eta-s$. Suppose further that the following conditions hold:
\begin{enumerate}
    \item There exists a function $K_{1} \in \mathcal{G}^0$ independent of $\eta$ such that
    $$
    \left|\mathbb{E} \left[\prod_{j=1}^{k} D_{\left(i_{j}\right)}(s)\right]-\mathbb{E} \left[\prod_{j=1}^{k} \widetilde{D}_{\left(i_{j}\right)}(s)\right]\right| \leq K_{1}(s)\eta^{\alpha+1}
    $$
    for $k=1,2, \ldots, \alpha$ and
    $$
    \mathbb{E} \left[\prod_{j=1}^{\alpha+1}\left|D_{\left(i_{j}\right)}(s)\right|\right] \leq K_{1}(s) \eta^{\alpha+1}
    $$
    for all $i_{j} \in\{1, \cdots, d\}$.
    \item For each integer $m \geq 1$, the $2m$-moment of $x_{t}$ is uniformly bounded with respect to $t$ and $\eta$,
    i.e. there exists a $K_{2} \in \mathcal{G}^0,$ independent of $\eta, t,$ such that
    $$\mathbb{E}\left[\left\|x_{t}^{s}\right\|_2^{2m}\right] \leq K_{2}(s),\quad\text { for all } t=0, \cdots, N.$$
\end{enumerate}

Then, for each $g \in \mathcal{G}^{\alpha+1},$ there exists a constant $C>0$ independent of $\eta$ such that
$$
\max _{t=0, \ldots, N}\left|\mathbb{E} g\left(x_{t}\right)-\mathbb{E} g\left(X_{t \eta}\right)\right| \leq C\eta^{\alpha}.
$$
\end{Lemma}

Note that the first condition in Lemma \ref{lemma:one_to_n} is about moment matching for one-step differences, which are studied in Lemma \ref{lemma:moment_D} and Lemma \ref{lemma:moment_C} under Assumptions \ref{ass:L} and \ref{ass:more_L}. However, Assumptions \ref{ass:L} and \ref{ass:more_L} are in general not sufficient to guarantee the second condition in Lemma \ref{lemma:one_to_n} on the uniform moment bound for the discrete-time process. Hence, we add the  the following assumption regarding the loss function $L$.

\begin{Assumption}\label{ass:more_more_L}
The loss function $L$ and its derivatives satisfy the following linear growth conditions: there exist some function $L:\R^d\to\R_+$, such that $\E_{x\sim\P_x}[L(x)^m]<\infty$ for any integer $m>1$, and for any $\theta\in\R^\theta$, 
$$\|\nabla_\theta L(\theta,x)\|_2\leq L(x)(1+\|\theta\|_2),\quad\|\nabla_{x\theta} L(\theta,x)\nabla_x L(\theta,x)\|_2\leq L(x)(1+\|\theta\|_2).$$
\end{Assumption}

\begin{Theorem}{(Approximation)}\label{thm:weak_conv}
    Assume Assumptions \ref{ass:L}, \ref{ass:more_L}, and \ref{ass:more_more_L}. Fix an arbitrary time horizon $T>0$ and take the learning rate $\eta \in(0,1 \wedge T)$ and set the number of iterations $N=\left\lfloor\frac{T}{\eta}\right\rfloor.$ Let $\{\theta_t\}_{t=0,1,\cdot,N}$ be the discrete-time adversarial training dynamic defined in \eqref{equ:one_step_theta}, and  $\{\Theta_t\}_{t\in[0,T]}$ be the continuous-time SDE dynamic defined in \eqref{equ:sde_dynamics}. Set $\Theta_0=\theta_0$. Then $\{\Theta_t\}_{t\in[0,T]}$ is an order-2 weak approximation of $\{\theta_t\}_{t=0,1,\cdot,N}$. That is,  for each $g \in \mathcal{G}^{3},$ there exists a constant $C>0$ independent of $\eta$ such that
    $$
    \max _{t=0, \ldots, N}\left|\mathbb{E} g\left(\theta_{t}\right)-\mathbb{E} g\left(\Theta_{t \eta}\right)\right| \leq C\eta^{2}.
    $$
\end{Theorem}

Theorem \ref{thm:weak_conv} states that the approximation error between the continuous-time SDE dynamic \eqref{equ:sde_dynamics} and the discrete-time training dynamic is in the order of $\Oc(\eta^2)$. Note that the approximation is in the sense of distribution of trained parameters, meaning this result holds for a class of neural networks 
from this distribution.
Note also the particular order of error bound can vary depending on the specific form of continuous-time SDE dynamics, and may be further improved. 

\subsection{Convergence Analysis via Invariant Measure}\label{sec:convergence}
Additionally, by studying the invariant measure of the SDE \eqref{equ:sde_dynamics}, and by adopting methodologies and notations from \cite{veretennikov1988bounds},
\cite{bianca2017existence}, and \cite{da2006introduction}, 
we have  the convergence of the SDE 
under the following assumption:
\begin{Assumption}\label{ass:conv_L}
$b_0,b_1$ and $\sigma$ defined in \eqref{equ:b0}, \eqref{equ:b1} and \eqref{equ:sigma} satisfy the following regularity conditions.
    \begin{enumerate}
        \item $b_0,b_1$ and $\sigma$ are Lipschitz continuous.
        
        \item There exist some positive real numbers $r$ and $R$, independent of $\eta$, such that for any $\theta\in\R^{d_\theta}$ with $\|\theta\|_2\geq R$,
        \begin{equation}\label{eqn:ergodic}
            \theta^T\left(b_1(\theta)+\eta b_2(\theta)\right)\leq -r\|\theta\|_2.
        \end{equation}
        
        \item There exists a constant $l>0$ such that for any $\theta\in\R^{d_\theta}$,
        \begin{equation}\label{eqn:parabolic}
            \theta^T\Sigma(\theta)\theta\geq l\|\theta\|_2^2.
        \end{equation}
    \end{enumerate}
\end{Assumption}

\begin{Theorem}{(Convergence)}\label{thm:conv_measure}
    Assume Assumption \ref{ass:L} and Assumption \ref{ass:conv_L}. The stochastic process $\{\Theta_t\}_{t\geq0}$ satisfied the SDE \eqref{equ:sde_dynamics} admits a unique invariant measure $\mu^*$, with an exponential convergence rate. More specifically, there exist a constant $\rho>0$ and a positive function $C:\R^{d_\theta}\to\R_+$, such that for any measurable set $\A\subset\R^{d_\theta}$,
    $$\left|\P\left[\Theta_t\in\A|\Theta_0=\theta\right]-\P_\mu(\A)\right|\leq C(\theta)e^{-\rho t}.$$
\end{Theorem}

Mathematically, Assumption \ref{ass:conv_L}.1  guarantees the existence of a unique (strong) solution to the SDE \eqref{equ:sde_dynamics}. Assumption \ref{ass:conv_L}.2 is   closely related to the recurrence property of the process \cite{bianca2017existence}. Assumption \ref{ass:conv_L}.3 requires  $\Sigma(\theta)$ to be uniformly elliptic, and is also known as the non-degenerate condition \cite{karatzas2014brownian}.
Both Assumptions \ref{ass:conv_L}.2 and \ref{ass:conv_L}.3 are standard to study invariant measures and ergodicity of SDEs, see for example, \cite{veretennikov1988bounds}, \cite{khasminskii2011stochastic} and \cite{hong2019invariant}.

These assumptions provide useful insight for
algorithm designs of SGD. 
For instance, Assumption \ref{ass:L} requires the boundedness of the loss function's gradient. In particular, the boundedness assumption explains analytically some well-known practices in adversarial training, including the introduction of various forms of gradient penalties; see for example, \cite{yan2018deepdefense} and \cite{farnia2018generalizable}.
Assumption \ref{ass:conv_L}.2  suggests that the loss function and training algorithm need to be designed such that there exist strong gradient signals to keep 
$\theta$ from being extremely large;
in practice, this requirement can be satisfied by adding the $l_2$-regularization to $\theta$. Hence, Theorem \ref{thm:conv_measure} justifies mathematically from a continuous-time viewpoint the widely-used $l_2$ regularization helps to stabilize the training process.

\section{Comparison Between SGD and Adversarial Training}\label{sec:SGD_VS_AL}

\subsection{Difference in their perspective continuous-time approximations}
To start, first recall that the discrete-time dynamic for SGD is 
\begin{equation}\label{equ:SGD}
    \theta^{s}_{t+1}=\theta^{s}_t - \frac{\eta}{B}\sum_{j=1}^B\nabla_\theta L(\theta^{s}_t, \widehat{x}_j),
\end{equation}
and its continuous-time SDE approximation (see for example, \cite{li2019stochastic}) is   
\begin{equation}\label{equ:sde_dynamics_SGD}
    d\Theta^{s}_t=(b^{s}_0(\Theta^{s}_t)+\eta b^{s}_1(\Theta^{s}_t))dt+\sigma^{s}(\Theta^{s}_t)dW_t
\end{equation}
with
\begin{align}
\begin{split}\label{equ:b0_s}
    b^s_0(\theta)&=-\nabla_\theta g(\theta),
\end{split}\\
\begin{split}\label{equ:b1_s}
    b^s_1(\theta)&=-\frac{1}{4}\nabla_\theta (\|\nabla_\theta g(\theta)\|^2),
\end{split}\\
\begin{split}\label{equ:sigma_s}
    \sigma^s(\theta)&=\sqrt{2\beta^{-1}}\left(\text{Var}_x(\nabla_\theta L(\theta,x))\right)^{1/2}.
\end{split}
\end{align}

\medskip
Comparing the continuous time approximations 
\eqref{equ:sde_dynamics_SGD} and \eqref{equ:sde_dynamics}, it is clear that \eqref{equ:sde_dynamics_SGD} can be viewed as a special case of \eqref{equ:sde_dynamics} by taking $K$ $=$ $0$. The term $b_0(\theta)$ in \eqref{equ:b0} differs from $b^{s}_0(\theta)$ in \eqref{equ:b0_s} by an additional correction term $K\beta^{-1}(\E[\|\nabla_x L(\theta,x)\|^2] - \|I(\theta)\|^2)$, due to adversarial perturbations. 
Meanwhile, the term $b_1(\theta)$ in \eqref{equ:b1} differs from $b^{s}_1(\theta)$ in \eqref{equ:b1_s} by two additional terms, $K\beta^{-1}(\E[\|\nabla_x L(\theta,x)\|^2] - \|I(\theta)\|^2)$ and $-\frac{K}{2}\nabla_\theta(\|I(\theta)\|^2)$, also caused by adversarial perturbations.

\subsection{Gradient Flow and Robustness of Adversarial Training}\label{subsec:robustness}

These differences between adversarial training and SGD enable us to explain analytically the robustness of adversarial training from a (new) gradient-flow viewpoint. 

Indeed, if  \eqref{equ:sde_dynamics} is  viewed (approximately) as the negative gradient flow with respect to the function $G(\theta)$ and if \eqref{equ:sde_dynamics_SGD} is viewed (approximately) as the negative gradient flow with respect to the expected loss function $g(\theta)$, then 
the extra term $\E[\|\nabla_x L(\theta,x)\|^2]$ for adversarial training  exactly reflects the sensitivity of the loss function $L$ under the perturbation of data $x$, contributing to the robustness of the trained model. 

In other words, the continuous-time SDE approximation offers a new perspective  to understand why using adversarial training results empirically in more robust models than simply applying SGD: adversarial training chooses $\theta$ to decrease the expected loss function $g(\theta)$ while simultaneously improving the  robustness according to the criterion $\E[\|\nabla_x L(\theta,x)\|^2]$, while SGD only focuses on decreasing the expected loss $g(\theta)$.

\subsection{Comparison through Examples}\label{subsec:exp_app}
The above analytical explanation can be corroborated in a numerical experiment with a logistic regression problem in Section \ref{subsec:log_reg}. Meanwhile,
more explicit comparison between the dynamic of adversarial training and  its SGD counterpart can be derived in the following
  linear model,  for which the corresponding SDE can be explicitly solved.

Let $H\in\mathbb{R}^{d\times d}$ be a symmetric positive definite matrix, with $0<\lambda_1\leq\lambda_2\leq\dots\leq\lambda_d$  its eigenvalues. Consider the loss function 
\begin{equation}\label{equ:quad_loss}
    L(\theta, x) = \frac{1}{2}(\theta-x)^T H (\theta-x) - \text{Tr}(H),
\end{equation}
and the data $x\sim\Nc(0,I)$. 
In this case, $$g(\theta):=\E_x[L(\theta, x)]=\frac{1}{2}\theta^T H \theta,$$
\begin{equation*}
    \nabla_xL(\theta, x)=-\nabla_\theta L(\theta, x)=H(x-\theta),\,\,\nabla_{x\theta}L(\theta,x)=H.
\end{equation*}
Hence, the SDE approximation \eqref{equ:sde_dynamics} for adversarial training applied to the model \eqref{equ:quad_loss} is
\begin{equation}\label{eqe:SDE_quad}
    d\Theta_t=-(H+(K+\frac{1}{2})\eta\cdot H^2)\Theta_t dt + \sqrt{2\beta^{-1}}HdW_t.
\end{equation}
This linear SDE \eqref{eqe:SDE_quad} is a multi-dimensional Ornstein-Uhlenbeck (OU) process and admits an explicit solution:
\begin{equation}\label{equ:SDE_quad_explicit}
    \Theta_t=e^{-\widehat{H}t}\Theta_0+ \sqrt{2\beta^{-1}} \int_0^t He^{-\widehat{H}(t-s)} dW_t, 
\end{equation}
with $\widehat{H}:=H+(K+\frac{1}{2})\eta\cdot H^2$.

By It$\hat{\text{o}}$'s isometry \cite{karatzas2014brownian}, we then deduce the dynamic of the objective function as
\begin{align}\label{equ:quad_loss_dynamic_AT}
    \E[g(\Theta_t)]=& \frac{1}{2}\Theta_0^T H e^{-2\widehat{H}t} \Theta_0+\frac{1}{\beta}\int_0^t\text{Tr}(H^3e^{-2\widehat{H}(t-s)}) ds\notag\\
    =& \frac{1}{2}\Theta_0^T H e^{-2\widehat{H}t} \Theta_0+\frac{1}{\beta}\int_0^t\sum_{i=1}^d \lambda_i^3 e^{-2(\lambda_i+(K+\frac{1}{2})\eta\lambda_i^2)(t-s)} ds\notag\\
    =& \frac{1}{2}\Theta_0^T H e^{-2\widehat{H}t} \Theta_0+\frac{1}{\beta}\sum_{i=1}^d\frac{\lambda_i^2\left(1-e^{-(2\lambda_i+(2K+1)\eta\lambda_i^2)t}\right)}{2+(2K+1)\eta\lambda_i} .
\end{align}
Meanwhile, the SDE approximation to the vanilla SGD for model \eqref{equ:quad_loss} has the following dynamic \cite{li2019stochastic}:
\begin{equation*}
d\Theta^{s}_t=-H\Theta^{s}_t dt + \sqrt{2\beta^{-1}}HdW_t,
\end{equation*}
with $\Theta^{s}_0=\Theta_0$, and
\begin{align}\label{equ:quad_loss_dynamic_SGD}
    &\E[g(\Theta^{s}_t)]=\frac{1}{2}\Theta_0^T H e^{-2Ht} \Theta_0+\beta^{-1}\sum_{i=1}^d \frac{\lambda_i^2}{2} \left(1-e^{-2\lambda_i t}\right).
\end{align}

As pointed out in \cite{li2019stochastic}, the first term in \eqref{equ:quad_loss_dynamic_SGD} decays exponentially with an asymptotic rate $2\lambda_d$, and the second term is induced by noise with its asymptotic value proportional to the learning rate $\eta$ as $\beta^{-1}=\frac{\eta}{2B}$. This is the well-known two-phase behavior of SGD under constant learning rate: an initial descent phase induced by the deterministic gradient flow and an eventual fluctuation phase dominated by the variance of the stochastic gradients. 

Now in adversarial training, the first term of the dynamic of the objective function \eqref{equ:quad_loss_dynamic_AT} also decays exponentially, but with a faster asymptotic rate $2\lambda_d+(2K+1)\eta\lambda_d^2$. This is because the gradient descent direction with respect to $\theta$ under this particular model \eqref{equ:quad_loss} coincides with the gradient ascent direction with respect to $x$, that is $\nabla_x L(\theta, x)=-\nabla_\theta L(\theta, x)$. Hence, the inner loop is  accelerating the convergence of $\theta$.  For the second term induced by the noise, its asymptotic value is also proportional to the learning rate $\beta^{-1}=\frac{\eta}{2B}$ when $\eta\to 0$.

\paragraph{Numerical testing} The explicit calculation for the model \eqref{equ:quad_loss}   enables us to verify numerically the second-order approximation result in Theorem \ref{thm:weak_conv}. 
To see this, take the model \eqref{equ:quad_loss} with a randomly generated 10-by-10 positive definite matrix $H$, and a random initial value $\theta_0$. The regularization (penalty) term $R(\delta)$ in \eqref{equ:adv_objective_new} is set to be $\|\delta\|_2^2$, along with other  parameters choices of  $B=20, K=5, T=2.0, \lambda=2.0$.

As suggested by Theorem \ref{thm:weak_conv},  the error with the test function is equal to the expected loss function: $f(\theta)=\frac{1}{2}\theta^TH\theta$.
The expectation of adversarial training dynamic $\E [f(\theta_{[T/\eta]})]$ is averaged over 1e5 runs, while the expectation of continuous-time SDE dynamic $\E [f(\Theta_{T})]$ is computed via the explicit formula \eqref{equ:quad_loss_dynamic_AT}.
\begin{figure}[H]
    \centering
    \includegraphics[width=0.5\linewidth]{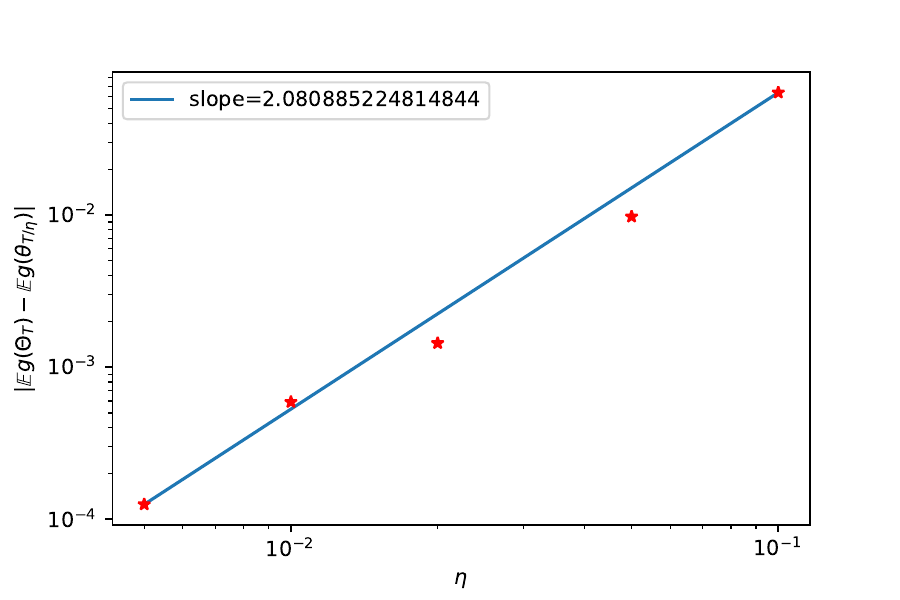}
    \caption{\centering Log-log plot of $|\mathbb{E}g(\Theta_T)-\mathbb{E}g(\theta_{T/\eta})|$ and $\eta$.}
    \label{fig:order}
\end{figure}
Figure \ref{fig:order} shows the Log-log plot of $|\mathbb{E}g(\Theta_T)-\mathbb{E}g(\theta_{T/\eta})|$ under different $\eta$. Theorem \ref{thm:weak_conv} suggests that the slope should be close to 2, which is consistent with the numerical result of 2.081.

%%%%%%%%%%%%%%%%%%%%%%%%%%%%%%%%%%%%%%%%%%%
%%%%%%%%%%%%%%%%%%%%%%%%%%%%%%%%%%%%%%%%%%%
%%%%%%%%%%%%%%%%%%%%%%%%%%%%%%%%%%%%%%%%%%%

%%%%%%%%%%%%%%%%%%%%%%%%%%%%%%%%%%%%%%%%%%%
%%%%%%%%%%%%%%%%%%%%%%%%%%%%%%%%%%%%%%%%%%%
\section{From Robust Optimization to Adversarial Training}\label{sec:RO_vs_AT}
In this section, we draw the connection between robust optimization \cite{ben2009robust} and adversarial training through a robust portfolio selection problem \cite{blanchet2021distributionally, data_driven}.

\paragraph{Robust portfolio selection problem} Consider a capital market consisting of $d$ assets whose yearly returns are captured by the random vector $\xi = [\xi_1, ..., \xi_d]^T \sim \mathbb{P}.$ The goal is to find a portfolio allocation vector $\phi = [\phi_1, ..., \phi_d]^T$ in the unit simplex $\mathbb{X} := \{\phi \in \mathbb{R}^d_+: \sum_{j=1}^d \phi_j = 1 \}.$ Since a portfolio $\phi$ invests a percentage $\phi_i$ of the available capital in asset $i$ for each $i=1,...,d,$ its return is $\langle \phi, \xi \rangle.$ 

If the full information of the return distribution $\mathbb{P}$ is given, the problem of finding the optimal portfolio allocation can be formulated as the following single-stage stochastic program:        
\begin{equation}\label{eqn:min}
    \min _{\phi \in \mathbb{X}}\left\{\mathbb{E}^{\mathbb{P}}[-\langle \phi, \xi\rangle]+\rho \mathrm{CVaR}_{\alpha}^\mathbb{P}(-\langle \phi, \xi\rangle)\right\}.
\end{equation}
The objective is a weighed sum of the mean and the conditional value-at-risk (CVaR) of the portfolio loss $- \langle \phi, \xi \rangle.$
%Intuitively, the %CVaR$_\alpha^\mathbb{P}(-%\langle \phi, \xi\rangle)$ %represents the mean of the %$\alpha \times 100\%$ worst %portfolio losses under the %distribution $\mathbb{P}$. 
According to the definition of CVaR in \cite{cVaR}, we can replace CVaR in (\ref{eqn:min}) by its formal definition and obtain an equivalent formulation:
\begin{align} \label{eqn:min2}
(\ref{eqn:min}) = &\min _{\phi \in \mathbb{X}}\left\{\mathbb{E}^{\mathbb{P}}[-\langle \phi, \xi\rangle]+\rho \min _{\beta \in \mathbb{R}} \mathbb{E}^{\mathbb{P}}\left[\beta+\frac{1}{\alpha} \max \{-\langle \phi, \xi\rangle-\beta, 0\}\right]\right\}\notag\\
= &\min _{\phi \in \mathbb{X}, \beta \in \mathbb{R}} \mathbb{E}^{\mathbb{P}}\left[-\langle \phi, \xi\rangle+\rho \left(\beta+\frac{1}{\alpha} \max \{-\langle \phi, \xi\rangle-\beta, 0\}\right)\right]\notag\\
= & \min _{\phi \in \mathbb{X}, \beta \in \mathbb{R}} \mathbb{E}^{\mathbb{P}}\left[-\langle \phi, \xi\rangle+\rho \beta+\frac{\rho}{\alpha}  \left(-\langle \phi, \xi\rangle-\beta\right)_+ \right].
\end{align}

In practice, however, instead of the full information of the return distribution $\mathbb{P}$, one may only have access to the empirical distribution of $\xi$, where one can collect $N$ samples $\hat{\xi}_i \in \mathbb{R}^d, i=1,...,N,$ and obtain $\hat{P}_N$ the empirical distribution of $\xi$ by  $\hat{P}_N = \sum_{i=1}^N \delta(\hat{\xi}_i).$ 
In this case, instead of \eqref{equ:adv_objective_new} where adversary attacks are added directly to data points,  distributional robust optimization (DRO) (see for instance \cite{blanchet2021distributionally, data_driven, pflug2007ambiguity})  considers perturbations on the entire empirical distribution under the Wasserstein metric. That is, DRO studies the portfolio allocation problem \eqref{eqn:min2} with the assumption that $\mathbb{P}$  is within a Wasserstein ball around the empirical distribution $\hat{\mathbb{P}}_N$, and its goal is to find the optimal portfolio allocation under the worst-case criterion:
\begin{equation}\label{eqn:minmax}
\begin{aligned}
    &\min_{\phi \in \mathbb{X}, \beta \in \mathbb{R}} 
    \max_{\mathbb{Q}}
    \mathbb{E}^{\mathbb{Q}}\left[-\langle \phi, \xi\rangle+\rho \beta+\frac{\rho}{\alpha}  \left(-\langle \phi, \xi\rangle-\beta\right)_+ \right], \\
    &\text{subject to } \mathcal{W}_c(\mathbb{Q}, \hat{\mathbb{P}}_N) \leq \epsilon.
\end{aligned}
\end{equation}
Here the Wasserstein distance between two probability measures $\mathbb{Q}_1$ and $\mathbb{Q}_2$ is 
\begin{align*}
\mathcal{W}_c(\mathbb{Q}_1, \mathbb{Q}_2) := \inf_{\Pi} \{& \int c(\xi_1, \xi_2)\Pi(d\xi_1, d\xi_2), \Pi \text{ is a joint distribution of } \\ 
& \xi_1 \text{ and } \xi_2 \text{ with marginals } \mathbb{Q}_1 \text{ and } \mathbb{Q}_2\},
\end{align*}
where $c: \mathbb{R}^d \times \mathbb{R}^d \rightarrow \mathbb{R}_+$ is a nonnegative lower semicontinuous function satisfying
$c(a, b) = 0$ if and only if $a = b$.

Moreover, via a strong duality argument, \cite{blanchet2019quantifying} reformulates 
(\ref{eqn:minmax}) as the following min-max problem:
%(See also \cite{sinha2017certifying}, %\cite{ren2022distributionally}).
%Therefore, 
\begin{equation}
\label{eqn:min_max_final}
\begin{aligned}
    \inf_{\lambda\geq 0, \phi, \beta}\lambda \epsilon +  \frac{1}{N}\sum_{i=1}^N \max_{z_i}\left( \rho \beta + \frac{\rho}{\alpha} (-\phi^T z_i - \beta)_+ - \phi^T z_i - \lambda c(z_i, \hat{\xi}_i)\right).
\end{aligned}
\end{equation}
%\solved{connect 6.40 with 2.2, what is zi - perturbed sample, param to optimize is lam, phi, beta. Corresponds to our al setting}

Clearly, (\ref{eqn:min_max_final}) is a special case of the general adversarial learning problem (\ref{equ:adv_objective_new}) in Section \ref{sec:setting}, where $(\lambda, \beta, \phi)$ is the parameter $\theta$ in (\ref{equ:adv_objective_new}), $\hat{\xi}_i$ corresponds to the data point $x_i$, and $z_i$ is the perturbed data point $x_i+\delta_i$.
 It is an optimization problem with a regularization term whose parameter is determined by $\epsilon$. This regularized optimization formulation is consistent with  \cite{blanchet2021distributionally} for mean-variance portfolio selection, \cite{blanchet2019robust} for Lasso, and \cite{blanchet2019robust} for logistic regression.  %The use of the Wasserstein ball naturally gives rise to a regularization term whose regularization parameter is determined by the radius of the perturbation ball.
 
 \eqref{eqn:min_max_final} can be solved by the  adversarial algorithm, with the objective function $J_{\text{port}}$ as
\begin{equation}\label{eqn:loss_func}
    J_{\text{port}}(\lambda,\beta,\phi,z_i; \hat{\xi_i}) = \lambda \epsilon +  \frac{1}{N}\sum_{i=1}^N \rho \beta + \frac{\rho}{\alpha} (-\phi^T z_i - \beta)_+ - \phi^T z_i - \lambda \|z_i - \hat{\xi}_i\|_1.
\end{equation}
The exact algorithm with gradient updates is detailed below in Algorithm \ref{algo:Adv_portfolio}.

\begin{algorithm}[!ht]
  \caption{\textbf{Adversarial Training Algorithm Applied to \eqref{eqn:min_max_final}}}
  \label{algo:Adv_portfolio}
\begin{algorithmic}[1]
    \STATE \textbf{Input}: loss function $J_{\text{port}}$ (\ref{eqn:loss_func}), 
    training set $\{\hat{\xi}_i\}_{i=1}^N$, mini-batch size $B$, training step $T$, inner loop step $K$, learning rates for outer and inner loops $\eta_{O}, \eta_{I}$.
    \STATE \textbf{Initialize}: $\lambda_0,\beta_0,\phi_0$.
    \FOR {$ 1 \leq t \leq T$}
        \STATE Sample a mini-batch of size $B$: $\{\hat{\xi}_{i_1},\dots,\hat{\xi}_{i_B}\}$. 
        \STATE Set $z^{(j)}_0=\hat{\xi}_{i_j},j=1,\dots,B$.
        \FOR {$ 1 \leq k \leq K$}
            \STATE \begin{align*}
                \Delta z^{(j)}_{k}&= -\frac{\rho}{\alpha} 1\{-\phi^T z^{(j)}_{k-1} - \beta \geq 0\} \phi 
    -\phi -\lambda ( 2\cdot 1_{z^{(j)}_{k-1} - \hat{\xi}_{i_j} > 0} -1 ) \\
                z^{(j)}_{k} &= z^{(j)}_{k-1} + \eta_I \Delta z^{(j)}_{k}
            \end{align*}
            %$\delta_k=\delta_{k-1}+\frac{\eta_{I}}{B}\sum_{j=1}^B\nabla_\delta J(\theta_{t-1},\widehat{x}_j,\delta_{k-1})$
        \ENDFOR
        \STATE $\lambda_t = \lambda_{t-1} - \eta_O (\epsilon - \frac{1}{N}\sum_{j=1}^N \|\hat{\xi}_{i_j} - z^{(j)}_K\|_1)$
        \STATE $\Delta \phi_t = \frac{1}{N} \sum_{j=1}^N \frac{\rho}{\alpha}(-z^{(j)}_K) 1\{ -\phi^T_{t-1} z^{(j)}_K - \beta_{t-1} \geq 0\} - z^{(j)}_K$
        \STATE $\Delta \beta_t = \rho - \frac{\rho}{\alpha}  \frac{1}{N} \sum_{j=1}^N 1\{ -\phi^T_{t-1} z^{(j)}_K - \beta_{t-1} \geq 0\}$
        \STATE $\phi_t = \phi_{t-1} - \eta_O \Delta \phi_t,\quad \beta_t=\beta_{t-1} - \eta_O \Delta \beta_t$
    \ENDFOR
\end{algorithmic}
\end{algorithm}

In Section \ref{subsec:portfolio}, we will apply Algorithm \ref{algo:Adv_portfolio} to the robust portfolio selection problem \eqref{eqn:min_max_final}. The perturbation power of the adversarial algorithm will be studied through experiments with different hyper-parameters, and the numerical results are shown to be consistent with the SDE approximation developed in Section \ref{sec:approx}.

%%%%%%%%%%%%%%%%%%%%%%%%%%%%%%%%%%%%%%%%%%%
%%%%%%%%%%%%%%%%%%%%%%%%%%%%%%%%%%%%%%%%%%%
\section{Proofs of Key Results}\label{sec:proof}

\paragraph{Proof of Lemma \ref{lemma:moment_C}}
The uniqueness of the solution to the SDE \eqref{equ:sde_dynamics} follows from the Lipschitz condition in Assumption \ref{ass:more_L} and the standard existence and uniqueness result for SDEs, see for instance Theorem 5.2.9 in \cite{karatzas2014brownian}.

To provide the moment estimates, let us first define the following operators for any test function $\psi\in\mathcal{G}^4$, with $\mathcal{G}^4$  the set of functions whose partial derivatives up to order four have at most polynomial growth:
\begin{align*}
    &\mathcal{L}_{1} \psi(\theta)=b_{0}(\theta)^{T} \nabla \psi(\theta),\\
    &\mathcal{L}_{2} \psi(\theta)=b_{1}(\theta)^{T} \nabla \psi(\theta) + \frac{1}{2B} \operatorname{Tr}\left(\Sigma(\theta) \nabla^{2} \psi(\theta)\right),\\
    &\mathcal{L}_{3} \psi(\theta)=\frac{1}{\sqrt{B}}\Sigma(\theta)^{1/2}\nabla \psi(\theta).
\end{align*}
Note that here $\mathcal{L}_{1} \psi(\theta)$ and $\mathcal{L}_{2} \psi(\theta)$ are real-valued functions on $\R^{d_\theta}$, while $\mathcal{L}_{3} \psi(\theta)$ is a vector-valued function from  $\R^{d_\theta}$ to $\R^{d_\theta}$.
By It$\hat{\text{o}}$'s formula, for any test function $\psi\in \mathcal{G}^4$,
$$
\begin{aligned}
\psi\left(\Theta_{\eta}\right)=& \psi(\Theta_0)+\int_{0}^{\eta} \Lc_1 \psi\left(\Theta_{s}\right) ds+\eta \int_{0}^{\eta} \Lc_2 \psi\left(\Theta_{s}\right) ds +\sqrt{\eta} \int_{0}^{\eta} \Lc_3 \psi\left(\Theta_{s}\right) dW_{s}.
\end{aligned}
$$
Applying the above formula to $\Lc_1 \psi$, $\Lc_2 \psi$ and $\Lc_1^2 \psi$ yields
$$
\begin{aligned}
\psi\left(\Theta_{\eta}\right)=&\,\psi(\Theta_0)+\eta \Lc_1 \psi(\Theta_0)+\eta^{2}\left(\frac{1}{2} \Lc_1^{2}+\Lc_2\right) \psi(\Theta_0) \\
&+\eta \int_{0}^{\eta} \int_{0}^{s}\left(\Lc_2 \Lc_1+\Lc_1 \Lc_2\right) \psi\left(\Theta_{v}\right) dvds + \int_{0}^{\eta} \int_{0}^{s} \int_{0}^{v} \Lc_1^{3} \psi\left(\Theta_{r}\right) drdvds \\
&+\eta^{2} \int_{0}^{\eta} \int_{0}^{s} \Lc_2^{2} \psi\left(\Theta_{v}\right) dvds + \eta \int_{0}^{\eta} \int_{0}^{s} \int_{0}^{v} \Lc_2 \Lc_1^{2} \psi\left(\Theta_{r}\right) drdvds \\
&+\sqrt{\eta} \int_{0}^{\eta} \Lc_3 \psi\left(\Theta_{s}\right) dW_{s} + \sqrt{\eta} \int_{0}^{\eta} \int_{0}^{s} \Lc_3 \Lc_1 \psi\left(\Theta_{v}\right) dW_{v}ds \\
&+\sqrt{\eta} \int_{0}^{\eta} \int_{0}^{s} \int_{0}^{v} \Lc_3 \Lc_1^{2} \psi\left(\Theta_{r}\right) dW_{r}dvds + \eta^{3 / 2} \int_{0}^{\eta} \int_{0}^{s} \Lc_3 \Lc_2 \psi\left(\Theta_{v}\right) dW_{v}ds.
\end{aligned}
$$
Taking expectations on both sides, all terms in the integral are either equal to zero or of order $\mathcal{O}\left(\eta^{3}\right)$.Since all the integrands have at most third order derivatives in $b_{0}, b_{1}, \sigma_{0}$ and fourth order derivatives in $\psi$, by the assumption that $b_{0}, b_{1}, \sigma_{0} \in \mathcal{G}^{3}$ and $\psi \in \mathcal{G}^{4}$, all the integrands belong to $\mathcal{G}^0$. Thus, the expectation of each integrand is bounded by 
\begin{equation}\label{equ:moment}
\kappa_{1}\left(1+\sup _{t \in[0, \eta]} \mathbb{E}\left|\Theta_{t}\right|^{2 \kappa_{2}}\right),
\end{equation}
for some $\kappa_{1}, \kappa_{2}>0$. By the Lipschitz condition in Assumption \ref{ass:more_L} and standard moment estimates for SDEs, (see again Theorem 5.2.9 in \cite{karatzas2014brownian}), \eqref{equ:moment} is finite. 

Meanwhile, the last four stochastic integrals involving $dW$ are martingales and their expectations are equal to zero. Therefore, the expectations of all the integrals are $\mathcal{O}\left(\eta^{3}\right)$, and
\begin{equation}\label{equ:general_ito}
    \psi\left(\Theta_{\eta}\right)=\,\psi(\Theta_0)+\eta \Lc_1 \psi(\Theta_0)+\eta^{2}\left(\frac{1}{2} \Lc_1^{2}+\Lc_2\right) \psi(\Theta_0) + \Oc(\eta^3)
\end{equation}
Therefore Lemma \ref{lemma:moment_C} follows by taking $\psi(\Theta_\eta)=\widetilde{D}_i,\widetilde{D}_i\widetilde{D}_j$, and $\widetilde{D}_i\widetilde{D}_j\widetilde{D}_k$ in \eqref{equ:general_ito}, respectively.

\paragraph{Proof of Theorem \ref{thm:weak_conv}}
It suffices to check that conditions in Lemma \ref{lemma:one_to_n} are satisfied by $\{\theta_t\}_{t=0,1,\cdot,N}$ and $\{\Theta_t\}_{t\in[0,T]}$ for $\alpha=2$.
    
    The first condition on one-step differences in Lemma \ref{lemma:one_to_n} holds by Lemma \ref{lemma:moment_D} and Lemma \ref{lemma:moment_C}. It remains to check the second condition on bounded moments.
    
    Recall that for any given $\{\widehat{x}_j\}_{j=1}^B$ sampled independently from $\P_x$, the update in adversarial training is
    \begin{align*}
    \theta_{t+1}&=\theta_t-\frac{\eta}{B}\sum_{j=1}^B\nabla_\theta L\left(\theta_{t},\widehat{x}_j\right)-\frac{K\eta^2}{B^2}\sum_{i,j=1}^B\nabla_{x\theta} L(\theta_{t},\widehat{x}_j)\nabla_x L(\theta_t,\widehat{x}_i)+\Oc(\eta^3)\\
    &:=\theta_t + \eta h\left(\theta_t,\{\widehat{x}_i\}_{i=1}^B,\eta\right),\text{ where}
\end{align*}
\begin{equation}
    h\left(\theta_t,\{\widehat{x}_i\}_{i=1}^B,\eta\right):=-\frac{1}{B}\sum_{j=1}^B\nabla_\theta L\left(\theta_{t},\widehat{x}_j\right)-\frac{K\eta}{B^2}\sum_{i,j=1}^B\nabla_{x\theta} L(\theta_{t},\widehat{x}_j)\nabla_x L(\theta_t,\widehat{x}_i)+\Oc(\eta^2).
\end{equation}
By Assumption \ref{ass:more_more_L}, the 2-norm of $h\left(\theta_t,\{\widehat{x}_i\}_{i=1}^B,\eta\right)$ is bounded by
\begin{align}
    \left\|h\left(\theta_t,\{\widehat{x}_i\}_{i=1}^B,\eta\right)\right\|_2&\leq\frac{1}{B}\sum_{j=1}^B\left\|\nabla_\theta L\left(\theta_{t},\widehat{x}_j\right)\right\|_2+\frac{K\eta}{B^2}\sum_{i,j=1}^B\left\|\nabla_{x\theta} L(\theta_{t},\widehat{x}_j)\nabla_x L(\theta_t,\widehat{x}_i)\right\|_2+\Oc(\eta^2)\notag\\
    &\leq\frac{1}{B}\sum_{j=1}^B  L(\widehat{x}_j)(1+\|\theta_t\|_2) + \frac{K\eta}{B^2}\sum_{i,j=1}^B L(\widehat{x}_j)(1+\|\theta_t\|_2)+\Oc(\eta^2)\notag\\
    &
    \leq\left(\frac{1}{B}\sum_{j=1}^B L(\widehat{x}_j) + \frac{K\eta}{B^2}\sum_{i,j=1}^BL(\widehat{x}_j) + 1\right)(1+\|\theta_t\|_2).\notag
\end{align}
Define the random variable
\begin{equation}\label{eqn:L_r}
    L_r:=\frac{1}{B}\sum_{j=1}^B L(\widehat{x}_j) + \frac{K\eta}{B^2}\sum_{i,j=1}^BL(\widehat{x}_j) + 1.
\end{equation}
By the independence of $\{\widehat{x}_j\}_{j=1}^B$ and the finite moment condition in Assumption \ref{ass:more_more_L}, we can conclude that $\E[L_r^m]<\infty$ for any $m>1$.

To simplify the notation, we now use $h_t$ to denote $h\left(\theta_t,\{\widehat{x}_i\}_{i=1}^B,\eta\right)$ in the remainder of the proof. Now for any integer $m>1$ and $t\geq0$, we have
$$
\left\|\theta_{t+1}\right\|_2^{m} \leq\left\|\theta_{t}\right\|_2^{m}+\sum_{l=1}^{m}\left(\begin{array}{c}
m \\
l
\end{array}\right)\left\|\theta_{t}\right\|_2^{m-l} \eta^{l}\left\|h_t\right\|_2^{l}.
$$

For $1 \leq l \leq m$, by Assumption \ref{ass:more_more_L} and the fact that $h_t$ is independent of $\theta_t$,
$$
\begin{aligned}
\mathbb{E}\left[\left\|\theta_{t}\right\|_2^{m-l}\left\|h_t\right\|_2^{l}\right] &=\mathbb{E}\left[\left\|\theta_{t}\right\|_2^{m-l} \mathbb{E}\left[\left\|h_t\right\|_2^{l} \mid \theta_{t}\right]\right] \\
&\leq\mathbb{E}\left[\left\|\theta_{t}\right\|_2^{m-l} \mathbb{E}\left[L_r^l(1+\|\theta_t\|_2)^l \mid \theta_{t}\right]\right]\\
&=\mathbb{E}\left[\left\|\theta_{t}\right\|_2^{m-l}\left(1+\|\theta_t\|_2\right)^l \mathbb{E}\left[L_r^l \mid \theta_{t}\right]\right]\\
&=\mathbb{E}\left[\left\|\theta_{t}\right\|_2^{m-l}\left(1+\|\theta_t\|_2\right)^l L_r^l\right]\\
&=\mathbb{E}\left[L_{r}^{l}\right]\mathbb{E}\left[\left\|\theta_{t}\right\|_2^{m-l}\left(1+\|\theta_t\|_2\right)^l\right] \\
&\leq M\mathbb{E}\left[L_{r}^{l}\right]\left(1+\E\left[\|\theta_t\|_2^m\right]\right).
\end{aligned}
$$
Here the constant $M$ in the last line is independent of $\eta$ and $t$, but may depend on the uniform $l^\text{th}$-moment bound for $l<m$, the existence of which can be justified by induction on $m$.
Hence, if we denote $a_{t}:=\mathbb{E}\left\|\theta_{t}\right\|_2^{m}$, we have
$a_{t+1} \leq(1+C \eta) a_{t}+C^{\prime} \eta,$
with $C, C^{\prime}>0$ independent of $\eta$ and $t$, which immediately implies
$ a_{t} \leq\left(a_{0}+C^{\prime} / C\right)(1+C \eta)^{t}-C^{\prime} / C 
 \leq\left(a_{0}+C^{\prime} / C\right) e^{(T / \eta) \log (1+C \eta)}-C^{\prime} / C 
 \leq\left(a_{0}+C^{\prime} / C\right) e^{C T}-C^{\prime} / C.
$
Therefore, the second condition on bounded moments in Lemma \ref{lemma:one_to_n} holds and the weak approximation result follows.

%%%%%%%%%%%%%%%%%%%%%%%%%%%%%%%%%%%%%%%%%%%%%%%%%%%%%%%%%%%%%%%%%%%%%%%%%%%%%%%%%%%%%%
%%%%%%%%%%%%%%%%%%%%%%%%%%%%%%%%%%%%%%%%%%%%%%%%%%%%%%%%%%%%%%%%%%%%%%%%%%%%%%%%%%%%%%
\section{Numerical Examples}\label{sec:numeric}
In this section, two numerical examples will be presented to further illustrate the theoretical results established in Section \ref{sec:approx}. More specifically, 
\begin{itemize}
\item the numerical experiment on logistic regression confirms  that adversarial training, when compared with the vanilla SGD, has the added advantage of improving the robustness of the model, consistent with the gradient flow perspective discussed in Section \ref{subsec:robustness};
    %\item  an experiment based on the linear %model introduced Section %\ref{subsec:exp_app} will demonstrate %numerically that the continuous-time SDE %dynamic is indeed a second-order %approximation to the discrete-time %training dynamic, as guaranteed in %Theorem \ref{thm:weak_conv};
    \item the robust portfolio optimization problem \eqref{eqn:min_max_final} is solved numerically using adversarial training. Impact of hyper-parameters on the robustness of training  will be discussed. % The advantage of adaptive learning rate over fixed rate in terms of convergence and stability will also be exhibited numerically, as studied in Section \ref{sec:learning_rate}.
    
\end{itemize}

\subsection{Logistic Regression and Robustness}\label{subsec:log_reg}
In this section a logistic regression model is adopted to  illustrate the robustness in adversarial training discussed in Section \ref{subsec:robustness}. 

We consider the randomly generated data $(x,y)$, where $y\sim\text{Bernoulli}(p)$ and  $x$ is sampled from a multivariate Gaussian distribution $\Nc(\mu_i,\Sigma_i)$ given $y=i$. We are to fit the data with a logistic regression model
%\begin{equation}\label{equ:log_reg}
$$\P(y=1|x)=\frac{e^{b+\theta^Tx}}{1+e^{b+\theta^Tx}},$$
%\end{equation}
with the cross-entropy loss function
%\begin{equation}\label{equ:log_reg_los%s}
   $$L(\theta,x,y)=\log(1+e^{b+\theta^Tx})-y(b+\theta^Tx).$$
%\end{equation}
Here $p\in(0,1), \mu_i\in\mathbb{R}^5, \Sigma_i\in\mathbb{R}^{5\times 5}$ are randomly generated. 

Meanwhile, we choose $R$ in \eqref{equ:adv_objective_new} to be the $l_2$ regularization, along with other parameters of $B=10, K=5, T=10.0, \lambda=2.0, \eta=0.005$. For simplicity, we fix the bias $b$ and only update $\theta$. 

\paragraph{Results} 

\begin{figure}[H]
    \centering
    \includegraphics[width=0.8\linewidth]{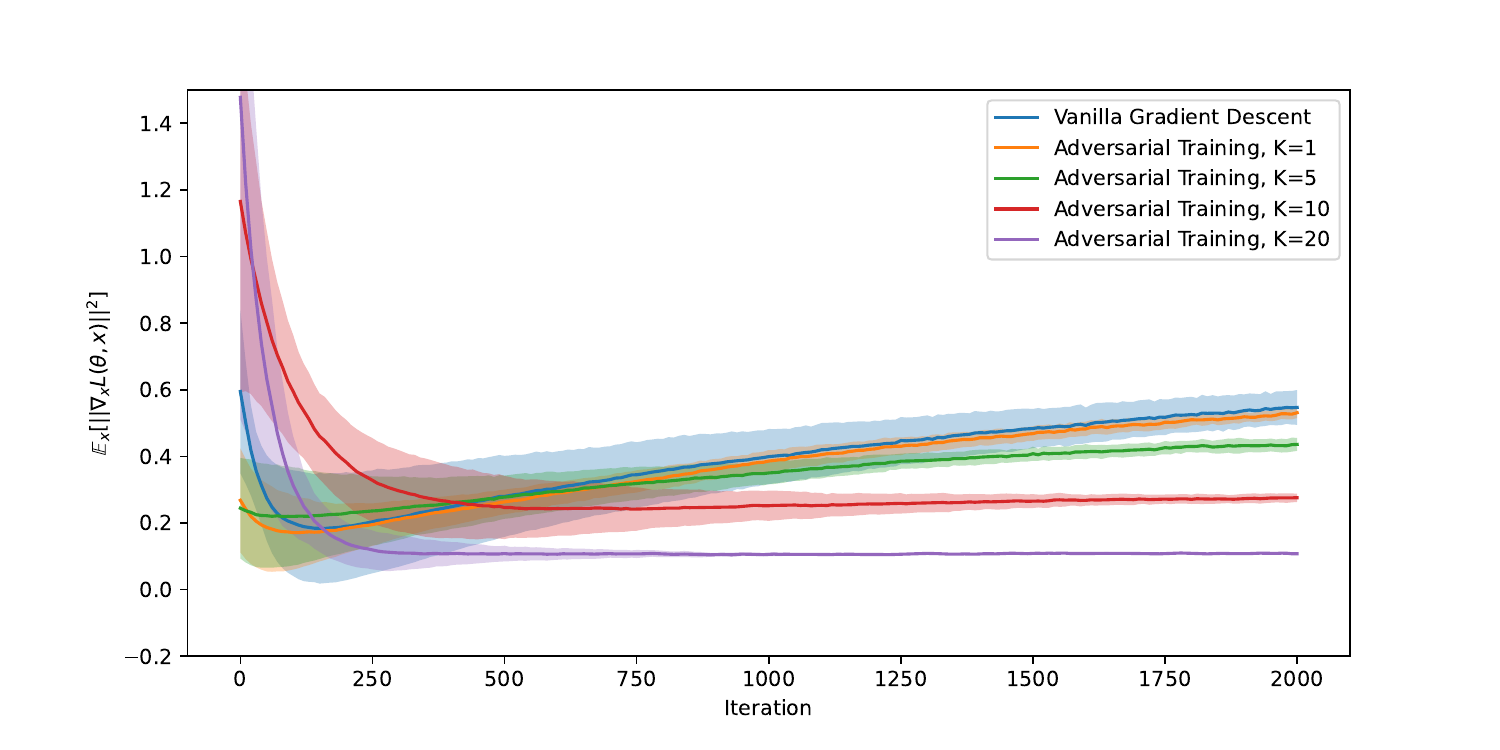}
    \caption{\centering\small Robustness criterion $\mathbb{E}_x[\|\nabla_x L(\theta, x,y)\|^2]$ of algorithms over training iterations.}
    \label{fig:logist_reg_robust}
\end{figure}

As suggested in Section \ref{subsec:robustness}, the models' robustness in terms of $\mathbb{E}_x[\|\nabla_x L(\theta, x,y)\|^2]$ for both stochastic gradient descent and adversarial learning algorithm over the training iterations is plotted in Figure \ref{fig:logist_reg_robust}, with means and standard deviations computed over 50 randomly initialized experiments.

Figure \ref{fig:logist_reg_robust} shows  how the number of perturbation steps $K$ affects the robustness of the model. When $K$ becomes larger (for example, $K=10$ or $20$), more adversarial perturbations are added to the training data, resulting in  more robust models  in terms of $\mathbb{E}_x[\|\nabla_x L(\theta, x,y)\|^2]$. In contrast, under vanilla gradient descent, it is observed that $\mathbb{E}_x[\|\nabla_x L(\theta, x,y)\|^2]$ first decreases then increases,  indicating that the model becomes increasingly sensitive hence less robust to adversarial perturbation as the training proceeds. 
This result is consistent with the gradient-flow viewpoint discussed in Section \ref{subsec:robustness}: 
updating via SGD only leads to  decrease in losses, while adversarial training leads to both improvement in robustness and reduction in losses.

Finally, note that in 50 randomly initialized experiments, SGD gets an average test accuracy of  84\%, while adversarial training with $K=10$ gets a slightly lower average test accuracy of 78\%. This finding is consistent with earlier empirical studies including \cite{athalye2018obfuscated} and \cite{zhang2020attack}, where adversarial training is shown to have a lower test accuracy on original samples in exchange for more robustness.

%xxixinxixNumerical results also illustrate that the adaptive learning rate discussed in Section \ref{sec:learning_rate} can improve the performance and stability of adversarial training in the portfolio selection problem.

\subsection{Robust Portfolio Selection}\label{subsec:portfolio}
In this experiment, we apply  Algorithm \ref{algo:Adv_portfolio} to the robust portfolio selection problem \eqref{eqn:min_max_final} discussed in Section \ref{sec:RO_vs_AT}. 
%The solution given by adversarial training (Algorithm \ref{algo:Adv_portfolio}) shows that in align with the SDE approximation developed in Section \ref{sec:approx}, the radius of the perturbation ball does not affect results of adversarial training. 
The perturbation power of the adversarial algorithm is studied through experiments with different hyper-parameters. 

\paragraph{Algorithm and numerical setup} Our experiments are based on a market with $d = 10$ assets. Following the similar setting of Section 7.2 in \cite{data_driven}, the return $\xi_i$ is decomposed into a systematic risk factor $\varphi \sim \mathcal{N}(0, 2\%)$ applied to all $d$ assets and an idiosyncratic risk factor $\zeta_i \sim \mathcal{N}(i\times 3\%, i \times 2.5\%).$
%\solved{move J out of the alogirhtm; change the format, centered equation with equation number}

%%%%%%%%%%%%%%%%%%%%%%%%%%%%%%%%%%%%%%%%%%%%%%%%%%%%%%%%%%%%%%%%%%%%%%%%%%%%%%%%%%%%%%
%%%%%%%%%%%%%%%%%%%%%%%%%%%%%%%%%%%%%%%%%%%%%%%%%%%%%%%%%%%%%%%%%%%%%%%%%%%%%%%%%%%%%%
\paragraph{Perturbation power of the adversarial learning algorithm}

\begin{figure}
     \centering
     \begin{subfigure}[b]{0.49\textwidth}
         \centering
         \includegraphics[width=\textwidth]{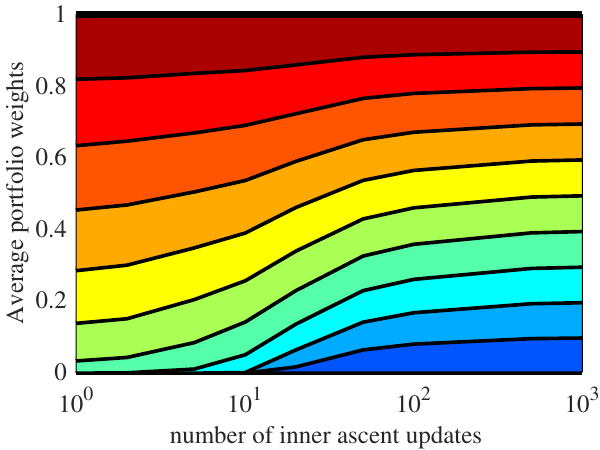}
         \caption{experiment with $K$}
         \label{fig:al_port1}
     \end{subfigure}
     \hfill
     \begin{subfigure}[b]{0.49\textwidth}
         \centering
         \includegraphics[width=\textwidth]{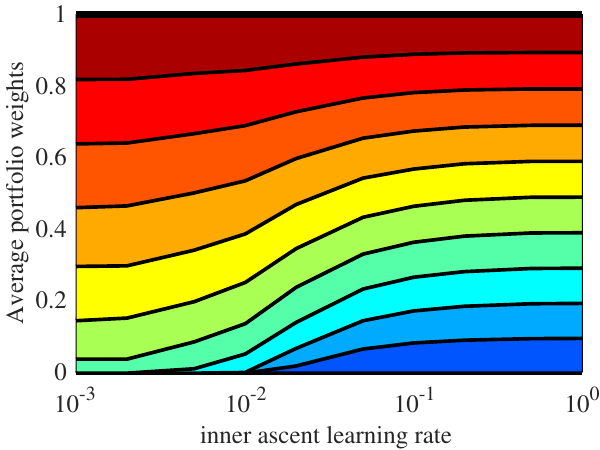}
         \caption{experiment with $\eta_I$}
         \label{fig:al_port2}
     \end{subfigure}
\caption{Portfolio composition by adversarial training as a function of  the number of inner updates $K$ in Figure \ref{fig:al_port1} and inner ascent learning rate $\eta_I$  in Figure \ref{fig:al_port2} averaged over 20 simulations with $N = 1000$ training samples and mini-batch size $B=100$, $\eta_O = 0.01$. For Figure \ref{fig:al_port1},  $\eta_I = 0.001$; For Figure \ref{fig:al_port2}, $K$=1.  }
\label{fig:al_port_al}
\end{figure}

The perturbation power of the adversarial algorithm is mainly determined by two hyper-parameters: number of inner ascent updates $K$ and inner ascent learning rate $\eta_I$. When $K$ and $\eta_I$ are small, the perturbations added to the original data are small; Consequently, assets with higher empirical returns are chosen with higher weights in the portfolio. On the other hand, for larger $K$ and $\eta_I$, the algorithm adds more perturbations to the original data, which will lead a more conservative portfolio selection. Indeed, as observed in Figure \ref{fig:al_port_al}, where Figure \ref{fig:al_port1} experiments with different $K$ values and Figure \ref{fig:al_port2} is with different $\eta_I$, when $K$ or $\eta_I$ increases, the outcome of Algorithm \ref{algo:Adv_portfolio} will converge to an equally-weighted portfolio.

\begin{figure}[H]
    \centering
 \begin{subfigure}[b]{0.49\textwidth}
         \centering
         \includegraphics[width=\textwidth]{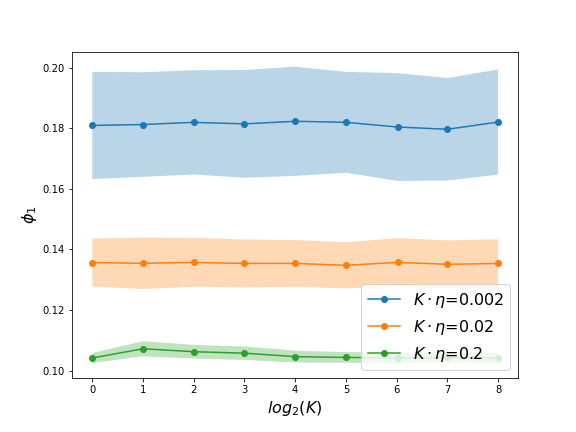}
         \caption{$\phi_1$ with different $K \cdot \eta_I$}
         \label{fig:K_eta_port1}
     \end{subfigure}
     \hfill
     \begin{subfigure}[b]{0.49\textwidth}
         \centering
         \includegraphics[width=\textwidth]{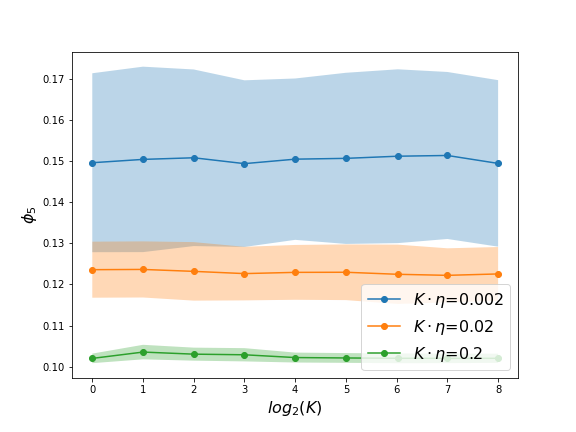}
         \caption{$\phi_5$ with different $K \cdot \eta_I$}
         \label{fig:K_eta_port5}
     \end{subfigure}   
     \caption{Portfolio composition by adversarial training 
     with different $K\cdot \eta_I$,
     averaged over 500 simulations with $N = 1000$ training samples and mini-batch size $B=100$.  }
    \label{fig:K_eta comparison}
\end{figure}

Another finding from Figure \ref{fig:al_port_al} is that, while keeping the product $K \cdot \eta_I$ unchanged (for example, $K=100, \eta_I=0.001$ in Figure \ref{fig:al_port1} v.s. $K=1, \eta_I=0.1$ in Figure \ref{fig:al_port2}), Algorithm \ref{algo:Adv_portfolio} yields similar portfolio selections, indicating that  $K \cdot \eta_I$ may be viewed as an ``effective'' perturbation power of the adversarial training. This observation  is consistent with the SDE approximation in Theorem \ref{thm:weak_conv} and \eqref{equ:inner_loop_delta}, which  clearly suggests   that $K$ and $\eta_I$ affect the perturbation through $K \cdot \eta_I$.
This conclusion is further demonstrated in Figure \ref{fig:K_eta comparison}, where three sets of experiments are conducted with $K \cdot \eta_I=0.2, 0.02$ and $0.002$. In each set of experiments, $K \cdot \eta_I$ is a fixed constant with $K$ varies from $2^0$ to $2^8$. The portfolio compositions from adversarial training under different choices of $K$ and $\eta_I$ are compared in terms of $\phi_1$ (the weight on the first asset) in Figure \ref{fig:K_eta_port1}, and $\phi_5$ (the weight on the fifth asset) in Figure \ref{fig:K_eta_port5}. Means and standard deviations of $\phi_1$ and $\phi_5$ are computed over 500 simulations. Figure \ref{fig:K_eta comparison} verifies that as long as $K\cdot \eta_I$ remains constant, adversarial training, Algorithm \ref{algo:Adv_portfolio}, produces similar portfolio compositions. Additionally, Figure \ref{fig:K_eta comparison} also indicates that larger adversarial perturbations (larger $K\cdot \eta_I$) lead to smaller standard deviations of the training outcomes, which is a clear improvement on the robustness.

%%%%%%%%%%%%%%%%%%%%%%%%%%%%%%%%%%%%%%%%%%%
%%%%%%%%%%%%%%%%%%%%%%%%%%%%%%%%%%%%%%%%%%%

\bibliographystyle{apalike}
\bibliography{refs}

\newpage

\end{document}